\documentclass{article}

\PassOptionsToPackage{numbers, compress}{natbib}
\usepackage[final]{neurips_2024}

\usepackage[colorlinks = true,
            linkcolor = BlueViolet,
            urlcolor  = BlueViolet,
            citecolor = BlueViolet,
            anchorcolor = BlueViolet]{hyperref}
\usepackage{url}            
\usepackage{booktabs}       
\usepackage{amsfonts}       
\usepackage{nicefrac}       
\usepackage{microtype}      
\usepackage[dvipsnames]{xcolor}
\usepackage{enumitem}

\usepackage{graphicx}
\usepackage{subfigure}
\usepackage{multirow}
\usepackage{multicol}
\usepackage{amsmath}
\usepackage{amssymb}
\usepackage{mathtools}
\usepackage{bm}
\usepackage{amsthm}
\usepackage{cleveref}
\usepackage{algorithmic}
\usepackage{algorithm}
\usepackage{tikz}
\usepackage{wrapfig}

\newcommand{\dd}{{\rm d}}

\title{\Large{Diffuser{\large Lite}: Towards Real-time Diffusion Planning}}

\author{%
    Zibin Dong$^1$\thanks{Contact me at \texttt{zibindong@outlook.com}}  ~~Jianye Hao$^1$\thanks{Correspondence to:
Jianye Hao (\texttt{jianye.hao@tju.edu.cn})}~~ Yifu Yuan$^1$ ~~Fei Ni$^1$ ~~Yitian Wang$^2$ ~~Pengyi Li$^1$ ~~Yan Zheng$^1$ \\
    {\small$^1$College of Intelligence and Computing, Tianjin University} \\ 
    {\small$^2$UC San Diego Jacobs School of Engineering} \\
}

\begin{document}

\maketitle

\begin{figure}[h]
\hsize=\textwidth
\vspace{-2em}
\centering
\includegraphics[width=0.98\textwidth]{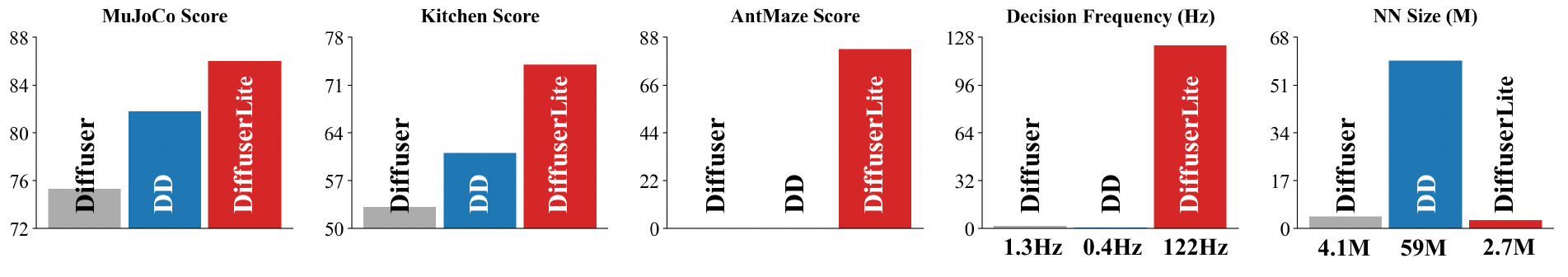}
\vspace{-12pt}
\caption{\small{\textbf{Performance overview.} We present Diffuser{\scriptsize Lite}, a lightweight framework that utilizes progressive refinement planning to reduce redundant information generation and achieves real-time diffusion planning. Diffuser{\scriptsize Lite} significantly outperforms predominant frameworks, Diffuser and DD, regarding scores, inference time, and model size on three popular D4RL benchmarks. The decision-making frequency of Diffuser{\scriptsize Lite} achieves \textbf{$\bm{122.2}$Hz}, which is \textbf{$\bm{112.7}$ times higher} than predominant frameworks.}}
\label{fig:1}
\end{figure}

\begin{tikzpicture}[remember picture,overlay]
\node[anchor=north west,inner sep=0pt,outer sep=0pt] at (0.5,8.8) 
{\includegraphics[width=0.8cm]{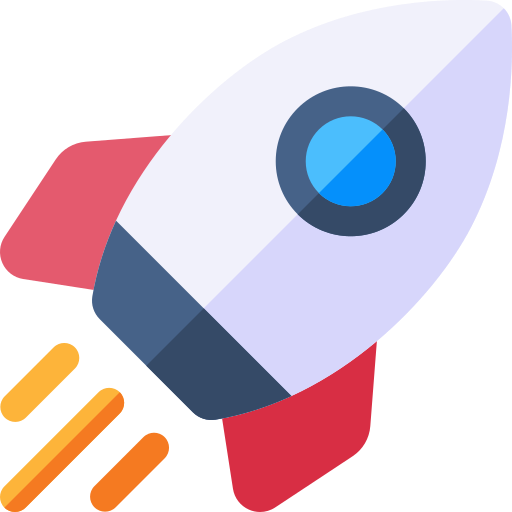}};  

\end{tikzpicture}


\vspace{-15pt}
\begin{abstract}
Diffusion planning has been recognized as an effective decision-making paradigm in various domains. The capability of generating high-quality long-horizon trajectories makes it a promising research direction. However, existing diffusion planning methods suffer from low decision-making frequencies due to the expensive iterative sampling cost. To alleviate this, we introduce \textbf{Diffuser{\footnotesize Lite}}, a super fast and lightweight diffusion planning framework, which employs a planning refinement process (PRP) to generate coarse-to-fine-grained trajectories, significantly reducing the modeling of redundant information and leading to notable increases in decision-making frequency. Our experimental results demonstrate that Diffuser{\footnotesize Lite} achieves a decision-making frequency of $122.2$Hz ($112.7$x faster than predominant frameworks) and reaches state-of-the-art performance on D4RL, Robomimic, and FinRL benchmarks. In addition, Diffuser{\footnotesize Lite} can also serve as a flexible plugin to increase the decision-making frequency of other diffusion planning algorithms, providing a structural design reference for future works. More details and visualizations are available at \href{https://diffuserlite.github.io/}{project website}.
\end{abstract}

\section{Introduction}

Diffusion models (DMs) are powerful generative models that demonstrate promising performance across various domains \citep{zhang2023controlnet, Ruiz2023dreambooth, Liu2023audioldm,kawar2023imagic}. Motivated by their remarkable capability in complex distribution modeling and conditional generation, researchers have developed a series of works applying diffusion models for decision-making tasks in recent years \citep{zhu2023diffusionsurvey}. DMs can play various roles in decision-making tasks, such as acting as planners to make better decisions from a long-term perspective \citep{janner2022diffuser, ajay2023dd, dong2023aligndiff, ni2023metadiffuser, du2023unipi}, serving as policies to support complex multimodal-distribution modeling \citep{pearce2023diffusionbc,wang2023dql, chi2023diffusionpolicy}, and working as data synthesizers to assist reinforcement learning (RL) training \citep{lu2023synther,yu2023rosie,he2023MTDiff}, etc. Among these roles, diffusion planning is the most widely applied paradigm \citep{zhu2023diffusionsurvey}. Unlike auto-regressive planning in previous model-based RL approaches \citep{hafner2019planet, Thuruthel2019loop, Hansen2022tdmpc}, diffusion planning avoids severe compounding errors by directly generating the entire trajectory rather than one-step transition \citep{zhu2023diffusionsurvey}. Also, its powerful conditional generation capability allows planning at the trajectory level without being limited to step-wise shortsightedness. The diffusion planning paradigm has achieved state-of-the-art (SOTA) performance in various offline RL tasks, including single-agent RL \citep{li2023hdmi}, multi-agent RL \citep{zhu2023madiff}, meta RL \citep{ni2023metadiffuser}, and more.

\begin{wrapfigure}{r}{7cm}
\centering
    \includegraphics[width=0.5\textwidth]{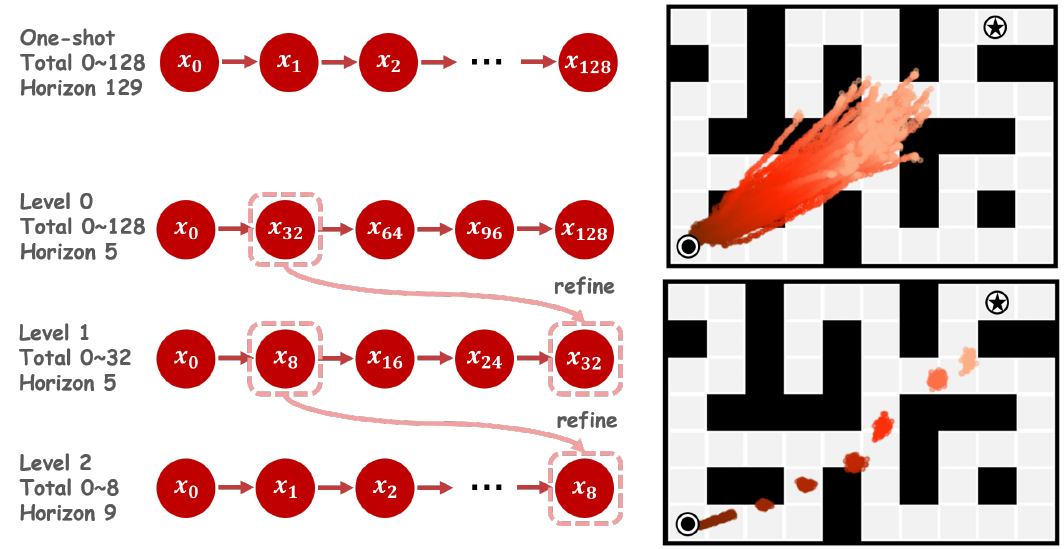}
    \vspace{-4pt}
    \caption{\small{\textbf{Comparison of one-shot planning (top) and PRP (down) on Antmaze.} The former directly generates plans with a temporal horizon of $129$. The latter consists of three coarse to fine-grained levels with temporal horizons of $0$-$128$, $0$-$32$, and $0$-$8$, and temporal jumps of $32$, $8$, and $1$, respectively. The visualization in the figure illustrates the x-y coordinates of $100$ plans. It shows that one-shot planning exhibits a significant amount of redundant information and a large search space. In contrast, PRP demonstrates better plan consistency and a smaller search space.}}
    \label{fig:2}
    \vspace{-10pt}
\end{wrapfigure}

One key issue that diffusion planning faces is the expensive iterative sampling cost. As depicted in \cref{fig:1}, the decision-making frequencies (number of actions inferred per second) of two predominant diffusion planning frameworks, Diffuser \citep{janner2022diffuser} and Decision Diffuser (DD) \citep{ajay2023dd}, are recorded as $1.3$Hz and $0.4$Hz, respectively. Such a low decision frequency fails to meet the requirements of numerous real-world applications, e.g. real-time robot control \citep{tai2018realtimerobot} and game AI \citep{pearce2023realtimegame}. The low decision frequency is primarily attributed to modeling a denoising process for a long-horizon trajectory distribution, which requires a heavy neural network backbone and multiple forward passes. A question may arise whether the generation of complete long-horizon trajectories is necessary for successful planning. Experimental results indicate that it is not, for the detailed trajectory information in long-horizon segments is highly redundant. As shown in \cref{fig:2}, a motivation example in Antmaze, the disparities between plans increase as the horizon grows, leading to poor consistency between plans in consecutive steps. Besides, in practice, agents often struggle to reach the planned distant state. These facts argue that while long-horizon planning helps improve foresight, it introduces redundant information in distinct parts. The details in closer parts are more crucial. Ignoring the modeling of these redundant parts in the diffusion planning process will significantly reduce the complexity of the trajectory distribution to be fitted, making it possible to build a fast and lightweight diffusion planning framework.

Motivated by these insights, we propose to build a plan refinement process (PRP) to speed up diffusion planning. First, we perform ``rough" planning, where jumpy planning is executed, only considering the states at intervals that are far apart and ignoring other individual states. Then, we refine a small portion of the plan, focusing on the steps closer to the current state. By doing so, we fill in the execution details between two states far apart, gradually refining the plan to the step level. This approach has three advantages: 1) It reduces the length of the sequences generated by the diffusion model, simplifying the complexity of the probability distribution to be fitted. 2) It significantly reduces the search space of plans, making it easier for the planner to find well-performed trajectories. 3) Since only the first action of each step is executed, rough planning of steps further away causes no noticeable performance drop.

\label{intro:claims}Diffusion planning with PRP, which we call Diffuser{\footnotesize Lite}, is simple, fast, and lightweight. Our experiments have demonstrated the effectiveness of PRP, significantly increasing decision-making frequency while achieving SOTA performance. Moreover, it can be easily adapted to other existing diffusion planning methods. In summary, our contributions are as follows:

\begin{itemize}
\item We introduce the plan refinement process (PRP) for coarse-to-fine-grained trajectory generation, reducing the modeling of redundant information.
\item We introduce Diffuser{\footnotesize Lite}, a lightweight diffusion planning framework, which significantly increases decision-making frequency by employing PRP.
\item Diffuser{\footnotesize Lite} is a simple and flexible plugin that can be easily combined with other diffusion planning algorithms. 
\item Diffuser{\footnotesize Lite} achieves a super high decision-making frequency ($122.2$Hz, $112.7$x faster than predominant frameworks) and SOTA performance on multiple benchmarks in D4RL.
\end{itemize}

\section{Preliminaries}
\textbf{Problem Setup:} Consider a system governed by discrete-time dynamics $o_{t+1}=f(o_t, a_t)$ at state $o_t$ given an action $a_t$. A trajectory $\bm x=[x_0,\cdots,x_{T-1}]$ can be either a sequence of states $x_t=o_t$ or state-action pairs $x_t=(o_t,a_t)$, where $T$ is the planning horizon. Each trajectory can be mapped to a property $\bm c$. Diffusion planning aims to find a trajectory that exhibits a property closest to the target:
\begin{equation}
    \bm x^*=\underset{\bm x}{\arg\min} ~d(\mathcal{C}(\bm x), \bm c_{\text{target}})
\end{equation}
where $d$ is a certain distance metric, $\mathcal{C}$ is a critic that maps a trajectory to the property it exhibits, and $c_{\text{target}}$ is the target property. The action to be executed $a_t$ is then extracted from the selected trajectory (\textit{state-action sequences}) or predicted by an inverse dynamic model $a_t = h(o_t, o_{t+1})$ (\textit{state-only sequences}). In the context of offline RL, it is a common choice to define the property as the corresponding cumulative reward $\mathcal C(\bm x)=\sum_{t=0}^{T-1}r(o_t, a_t)$ in previous works \citep{janner2022diffuser, ajay2023dd}.

\textbf{Diffusion Models} assume an unknown trajectory distribution $q_0(\bm x_0)$, DMs define a forward process $\{x_s\}_{s\in[0,S]}$ with $S>0$. Starting with $\bm x_0$, previous work \citep{kingma2021on} proved that one can obtain any $\bm x_s$ by solving the following stochastic differential equation (SDE):

\begin{equation} \label{eq:diffusion_forward_process}
    \dd\bm x_s=f(s)\bm x_s \dd s+g(s) \dd\bm w_s, ~\bm x_0\sim q_0(\bm x_0)
\end{equation}

where $\bm w_s$ is the standard Wiener process, and $f(s)=\frac{\dd\log\alpha_s}{\dd s}$, $g^2(s)=\frac{\dd\sigma_s^2}{\dd s}-2\sigma_s^2\frac{\dd\log\alpha_s}{\dd s}$. Values of $\alpha_s, \sigma_s \in \mathbb R^+$ depend on the noise schedule but keep the \textit{signal-to-noise-ratio} (SNR) $\alpha^2_s/\sigma^2_s$ strictly decreasing \citep{kingma2021on}. While this SDE transforms $q_0(\bm x_0)$ into a noise distribution $q_S(\bm x_S)=\mathcal{N}(\bm 0,\bm I)$, one can reconstruct trajectories from the noise by solving the reverse process of \cref{eq:diffusion_forward_process}. Previous work \citep{song2021scorebased} proved that solving its associated \textit{probability flow ODE} can support faster sampling:

\begin{equation}\label{eq:diff_ODE}
    \frac{\dd \bm x_s}{\dd s}=f(s)\bm x_s-\frac{1}{2}g^2(s)\nabla_{\bm x}\log q_s(\bm x_s), ~\bm x_S\sim q_S(\bm x_S)
\end{equation}

in which \textit{score function} $\nabla_{\bm x}\log q_s(\bm x_s)$ is the only unknown term and estimated by a neural network $-\bm\epsilon_\theta(\bm x_s,s)/\sigma_s$ in practice. The parameter $\theta$ is optimized by minimizing the following objective:
\begin{equation}
    \mathcal{L}(\theta)=\mathbb{E}_{q_0(\bm x_0),q(\bm\epsilon),s}[||\bm\epsilon_\theta(\bm x_s,s)-\bm\epsilon||^2_2]
\end{equation}
where $\bm\epsilon\sim q(\bm\epsilon)=\mathcal{N}(\bm 0,\bm I)$, $\bm x_s=\alpha_s\bm x_0 + \sigma_s\bm\epsilon$. Various ODE solvers can be employed to solve \Cref{eq:diff_ODE}, such as the Euler solver \citep{Atkinson1989eularsolver}, RK45 solver \citep{Dormand1980rk45}, DPM solver \citep{lu2022dpm}, etc.

\textbf{Conditional Sampling} helps to generate trajectories exhibiting certain properties in a priori. There are two main approaches: classifier-guidance (CG) \citep{dhariwal2021CG} and classifier-free-guidance (CFG) \citep{ho2021CFG}. CG requires an additional classifier $\log p_\phi(\bm c|\bm x_s, s)$ to predict the log probability that a noisy trajectory $\bm x_s$ exhibits a given property $\bm c$. The gradients from this classifier are then used to guide the solver:
\begin{equation}\label{eq:CG}
    \tilde{\bm\epsilon_\theta}(\bm x_s,s,\bm c):=\bm\epsilon_\theta(\bm x_s,s)-w\cdot\sigma_s\nabla_{\bm x_s}\log p_\phi(\bm c|\bm x_s,s)
\end{equation}
CFG does not require an additional classifier but uses a conditional noise predictor to guide the solver.
\begin{equation}\label{eq:CFG}
    \tilde{\bm\epsilon_\theta}(\bm x_s,s,\bm c):=w\cdot\bm\epsilon_\theta(\bm x_s,s,\bm c)+(1-w)\cdot\bm\epsilon_\theta(\bm x_s,s)
\end{equation}
Increasing the value of guidance strength $w$ leads to more property-aligned generation, but decreases the legality of the generated trajectories \citep{dong2023aligndiff}.


\begin{figure}
    \flushright
    \includegraphics[width=0.94\textwidth]{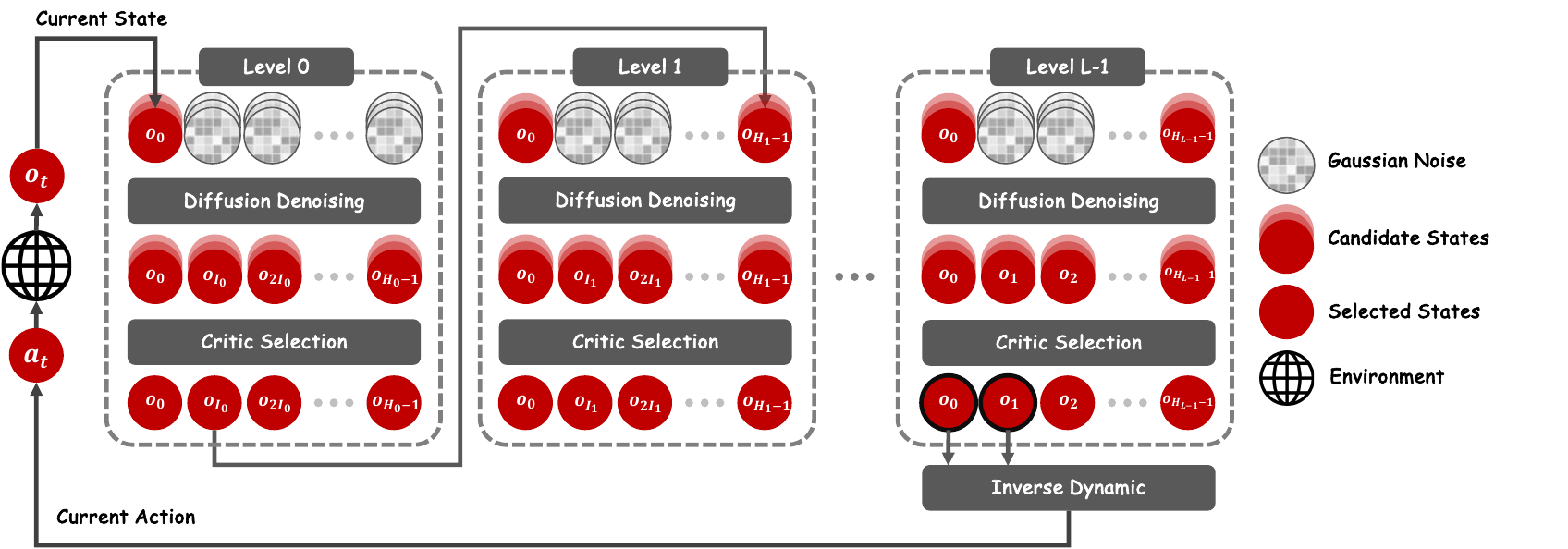}
    \vspace{-5pt}
    \caption{\small{\textbf{Overview of Diffuser{\scriptsize Lite}.} Observing the current state $o_t$, level $0$ of Diffuser{\scriptsize Lite} fixes $o_t$ as $o_0$ and generates multiple candidate trajectories. A critic is then used to select the optimal one, in which $o_{I_0}$ is then passed to the next level as its terminal $o_{H_1-1}$. The plan refinement process continues iteratively until the last level with a temporal jump of $I_{L-1}=1$. Finally, the action $a_t$ to be executed is extracted using an inverse dynamic model $a_t=h(o_0,o_1)$.}}
    \label{fig:3}
    \vspace{-5pt}
\end{figure}

\section{Efficient Planning via Refinement}

Diffuser \citep{janner2022diffuser} and DD \citep{ajay2023dd} are two pioneering frameworks, upon which a vast amount of diffusion planning works has been built \citep{liang2023adaptdiffuser, lee2023RGG, zhou2023replandiffuser}. Although the design details differ, they can be unified into one paradigm. In the inference step, multiple candidate trajectories are first conditionally sampled using the diffusion model. Then, a critic is used to select the optimal one that exhibits the closest property to the target. Finally, the action to be executed is extracted. This paradigm relies on multiple forwarding complex neural networks, resulting in extremely low decision-making frequencies (typically $1$-$10$Hz, or even less than $1$Hz), severely hindering its real-world deployment.

The fundamental reason is the requirement for highly complex neural networks to model the complex long-horizon trajectory distributions \citep{zhu2023diffusionsurvey}. Although some works have explored using advanced ODE solvers to reduce the sampling steps to around $5$ \citep{dong2023aligndiff}, the time consumption of network forwarding is still unacceptable. However, we notice that \textit{ignoring some redundant distant parts of the generated plan can be a possible solution} to address the issue. As shown in \cref{fig:2}, the disparities between plans increase as the horizon grows, leading to poor consistency between the plans selected in consecutive steps. Besides, agents often struggle to reach the distant states given by a plan in practice. Both facts indicate that terms in distant parts of a plan become increasingly redundant, whereas the closer parts are more crucial. 

Regarding these findings, we aim to develop a progressive refinement planning (PRP) process. This process initially plans a rough trajectory consisting of only key points spaced at equal intervals and then progressively refines the first interval by generative interpolating. It is worth noting that this is orthogonal to methods of identifying key points based on task semantic information \citep{li2023hdmi}. Specifically, our proposed PRP consists of $L$ planning levels. At each level $l\in \{0,1,\cdots, L-1\}$, starting from the known first term $x_0$, DMs plan rough trajectories $\bm x_{0:H_l:I_l}$ with temporal horizon $H_l$ and temporal jump $I_l$. Then, the planned first key point $x_{I_l}$ is passed to the next level as its terminal:
\begin{equation}
    \begin{aligned}
    \bm x_{0:H_l:I_l}:&=\left[ x_0,x_{I_l},x_{2I_l}\cdots,x_{H_l-1} \right] \\
    \bm x_{0:H_{l+1}:I_{l+1}}:&=\left[ x_0,x_{I_{l+1}},x_{2I_{l+1}},\cdots,x_{H_{l+1}-1} \right] \\
    x_{I_l}&=x_{H_{l+1}-1}
\end{aligned}
\end{equation}
By this design, only the first planned intervals are refined in the next level, and the other redundant details are all ignored, resulting in a coarse-to-fine generation process. The progressive refinement continues until the last level to extract an action. To support conditional sampling for each level, we define the property of a rough trajectory $\bm x_{0:H_l:I_l}$ as the property expectation over the distribution of all its completed trajectories $\mathcal X(\bm x_{0:H_l:I_l})$:
\begin{equation}
    \mathcal{C}(\bm x_{0:H_l:I_l}):=\mathbb E_{\bm x\sim\mathcal{X}(\bm x_{0:H_l:I_l})}[\mathcal{C}(\bm x)]
\end{equation}
PRP ensures that long-term planning maintains foresight while alleviating the burden of modeling redundant information. As a result, it greatly contributes to reducing model size and improving planning efficiency:

\textbf{Simplifying the fitted distribution of DMs.} The absence of redundant details in PRP allows for a significant reduction in the complexity of the fitted distribution at each level. This reduction in complexity enables us to utilize a lighter neural network backbone, shorter network input sequence lengths, and a reduced number of denoising steps.

\textbf{Reducing the plan-search space.} Key points generated at former levels often sufficiently reflect the quality of the entire trajectory, which allows the planner to focus more on finding distant key points and planning actions for the immediate steps, reducing search space and complexity.

\section{A Lite Architecture for Real-time Diffusion Planning}\label{sec:4}
Employing PRP results in a new lightweight architecture for diffusion planning, which we refer to as Diffuser{\footnotesize Lite}. Diffuser{\footnotesize Lite} can reduce the complexity of the fit distribution and significantly increase the decision-making frequency, achieving $122$Hz on average for the need of real-time control. We present the architecture overview in \cref{fig:3}, provide pseudocode for both training and inference in \cref{alg:training} and \cref{alg:inference}, and discuss detailed design choices in this section.

\textbf{Diffusion model for each level:} We train $L$ diffusion models for all levels to generate state-only sequences. We employ DiT \citep{Peebles2022DiT} as the noise predictor backbone, instead of the more commonly used UNet \citep{ronneberger2015unet} due to the significantly reduced length of the generated sequences in each level (typically around $5$). This eliminates the need for $1$D convolution to extract local temporal features. To adapt the DiT backbone for temporal generation, we make minimal structural adjustments following \citep{dong2023aligndiff}. For conditional sampling, we utilize CFG instead of CG, as the slow gradient computation process of CG reduces the frequency of decision-making. During the training phase, at each gradient step, we sample a batch of $H_0$-length trajectories, slice each of them into $L$ sub-trajectories $[\bm x_{0:H_{0}:1},\cdots,\bm x_{0:H_{L-1}:1}]$, evaluate their properties $\mathcal{C}(\bm x_{0:H_l:1})$ as an estimation of $\mathcal{C}(\bm x_{0:H_l:I_l})$ for condition at each level, and then slice the sub-trajectories into evenly spaced training samples $\bm x_{0:H_l:I_l}$ for training the diffusion models. During the inference phase, as shown in \cref{fig:3}, diffusion models generate multiple candidate plans level by level from 0 to $L-1$, and the optimal one is selected by the critic $\mathcal C$.

\textbf{Critic design:} The Critic $\mathcal C$ in Diffuser{\footnotesize Lite} plays two important roles: providing generation conditions during the diffusion training process and selecting the optimal plan from the candidates generated by the diffusion model during inference. In the context of Offline RL, both Diffuser and DD adopt the cumulative reward of a trajectory as the condition:
\begin{equation}\label{eq:cond_r}
    \mathcal{C}(\bm x)=\sum_{t=0}^{H-1}r(o_t,a_t),
\end{equation}
where $H$ is the temporal horizon. This design allows rewards from the offline RL dataset to be utilized as a ground-truth critic for acquiring generation conditions during training. During inference, an additional trained reward function is required to serve as the critic. The critic then helps select the plan that maximizes the cumulative reward, as depicted in the lower part of \cref{fig:3}. However, this design poses challenges in tasks with sparse rewards, as it can confuse diffusion models when distinguishing better-performing trajectories, especially for short-horizon plans. To address this challenge, we introduce an option to use the sum of discounted rewards and the value of the last state as an additional property design:
\begin{equation}\label{eq:cond_v}
    \mathcal{C}(\bm x)=\sum_{t=0}^{H-2}\gamma^tr(o_t,a_t)+\gamma^{H-1}V(o_{H-1}),
\end{equation}
where $V(o_t)=\max\mathbb E_{\pi}[\sum_{\tau=t}^{\infty} \gamma^{\tau-t}r_\tau]$ represents the optimal value function \citep{sutton1998RLbook} and can be estimated by a neural network through various offline RL methods. In the context of other domains, properties can be flexibly designed as needed, as long as the critic $\mathcal C$ can evaluate trajectories of variable lengths. It is worth noting that previous diffusion planning algorithms widely support this flexibility. In addition, it is even possible to skip the critic selection during inference, which is equivalent to using a uniform critic, as used in DD.

\textbf{Action extraction:} After obtaining the optimal trajectory from the last level through critic selection, we utilize an additional inverse dynamic model $a_t=h(o_t,o_{t+1})$ to extract the action to be executed. This approach is suggested in \citep{ajay2023dd}.

\textbf{Further speedup with rectified flow:}\label{sec:rf} Diffuser{\footnotesize Lite} aims to achieve real-time diffusion planning to support its application in real-world scenarios. Therefore, we introduce \textit{Rectified flow} \citep{liu2023rectifiedflow} for further increasing the decision-making frequency. Rectified flow, an ODE on the time interval $[0,1]$, causalizes the paths of linear interpolation between two distributions. If we define the two distributions as trajectory distribution and standard Gaussian, we can directly replace the Diffusion ODE with rectified flow to achieve the same functionality. The most significant difference is that rectified flow learns a straight-line flow and can continuously straighten the ODE through \textit{reflow}. This straightness property allows for consistent and stable gradients throughout the flow, enabling the generation of trajectories with very few sampling steps (in our experiments, we found that one-step sampling is sufficient to produce good results). We consider rectified flow an optional backbone for cases that prioritize decision frequency. In \cref{append:rectifiedflow}, we offer comprehensive explanations of trajectory generation and training with rectified flow.

\begin{wrapfigure}{r}{7cm}
    \centering
    \includegraphics[width=0.46\textwidth]{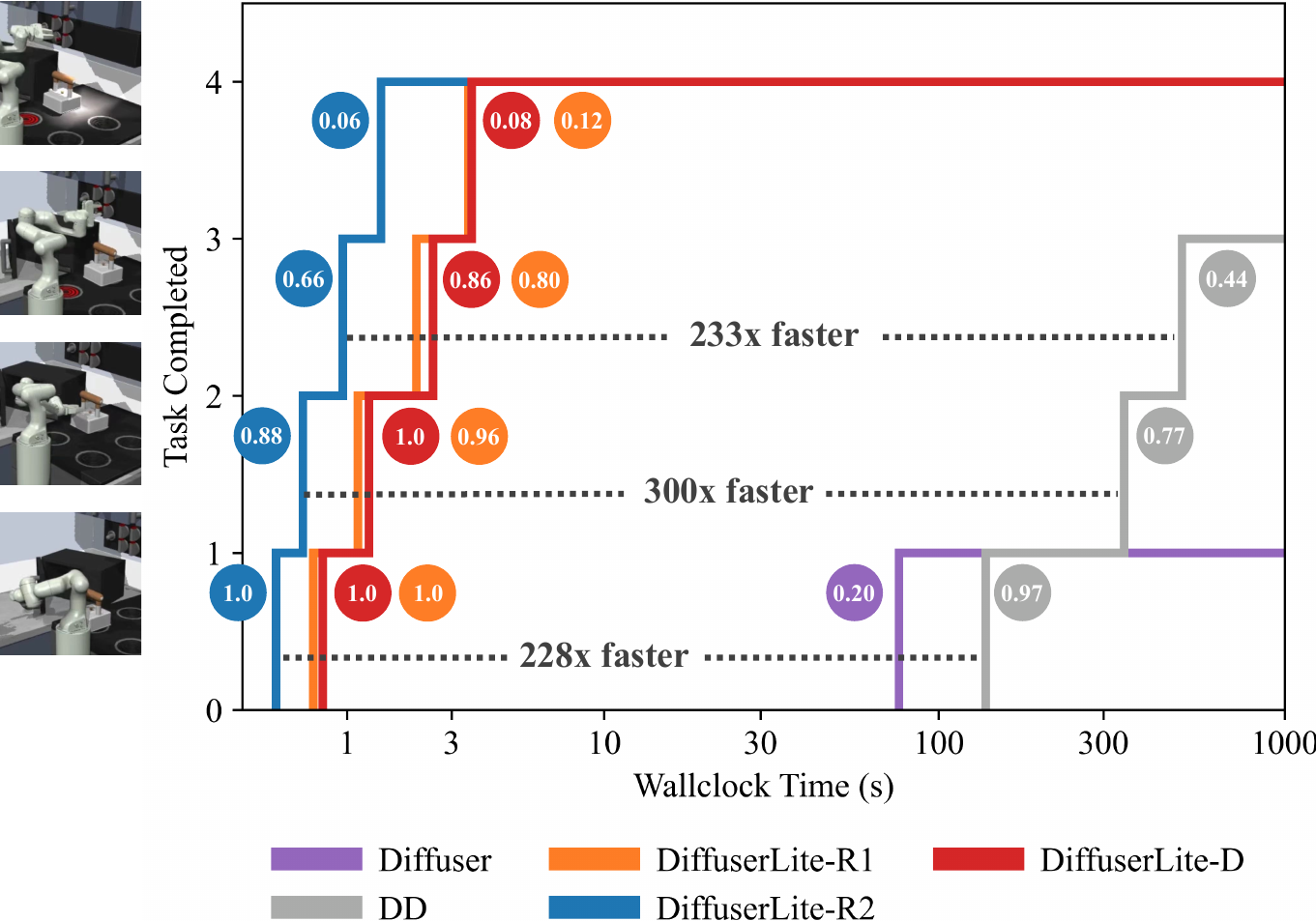}
    \vspace{-5pt}
    \caption{\small{\textbf{Runtime and performance comparison in FrankaKitchen.} The y-axis represents the number of completed tasks (maximum of $4$), and the x-axis represents the required wall-clock time. Task success rates are presented in colored circles. All results are averaged over 250 rollouts. Diffuser{\scriptsize Lite} demonstrates significant advantages in both wallclock time and success rate.}}
    \label{fig:4}
    \vspace{-20pt}
\end{wrapfigure}

\vspace{-9pt}
\section{Experiments}

\begin{table*}
\caption{\small{\textbf{Time consumption per step and frequency in D4RL.} All results are obtained over 5 random seeds. Diffuser{\scriptsize Lite} achieves an average decision frequency of about 122Hz on the R2 backbone and 81Hz averaging over three variations, which meets the requirements of real-time inference. Since the code for HDMI is not open-source, we make every effort to reproduce HDMI with the original settings to test its runtime cost and thus mark its results with underlines.}}
\label{tab:runtime}
\vskip 0.0in
\begin{center}
\begin{small}
\scalebox{0.83}{
\begin{tabular}{lccccccc}
\toprule
\textbf{Environment} & \textbf{Metric} & \textbf{Diffuser} & \textbf{DD} & \underline{\textbf{HDMI}} & \textbf{Diffuser{\scriptsize Lite}-D} & \textbf{Diffuser{\scriptsize Lite}-R1} & \textbf{Diffuser{\scriptsize Lite}-R2} \\ 
\midrule


\multirow{2}{*}{MuJoCo} 
& Runtime (s) & $0.665$ & $2.142$ & \underline{$0.405$} & $0.015$ & $0.013$ & $0.005$ \\
& Frequency (Hz) & $1.5$ & $0.47$ & \underline{$2.5$} & $68.2$ & $79.7$ & $200.7$ \\
\midrule 
\multirow{2}{*}{Kitchen} 
& Runtime (s) & $0.790$ & $2.573$ & \underline{$0.407$} & $0.017$ & $0.015$ & $0.010$ \\
& Frequency (Hz) & $1.3$ & $0.4$ & \underline{$2.5$} & $58.7$ & $66.0$ & $103.2$ \\
\midrule
\multirow{2}{*}{Antmaze} 
& Runtime (s) & $0.791$ & $2.591$ & \underline{$0.410$} & $0.027$ & $0.015$ & $0.010$ \\
& Frequency (Hz) & $1.3$ & $0.39$ & \underline{$2.4$} & $37.3$ & $65.7$ & $101.7$ \\
\midrule
\multicolumn{2}{c}{\textbf{Average Runtime (s)}} &
$0.749$ & $2.435$ & \underline{$0.407$} & $\bm{0.020}$ & $\bm{0.014}$ & $\bm{0.008}$ \\
\multicolumn{2}{c}{\textbf{Average Frequency (Hz)}} &
$1.34$ & $0.41$ & \underline{$2.46$} & $\bm{51.26}$ & $\bm{69.90}$ & $\bm{122.44}$ \\

\bottomrule
\end{tabular}}
\end{small}
\end{center}
\vspace{-15pt}
\end{table*}

We explored the performance of the Diffuser{\footnotesize{Lite}} on various tasks on D4RL, Robomimic, and FinRL \citep{fu2020d4rl, robomimic, neorl}, and aimed to answer the following research questions (RQs): 1) To what extent can Diffuser{\footnotesize{Lite}} reduce the \textbf{runtime cost}? 2) How is the \textbf{performance} of Diffuser{\footnotesize{Lite}} on offline RL tasks? 3) Can Diffuser{\footnotesize{Lite}} serve as a \textbf{flexible plugin} for other diffusion planning algorithms? 4) Can we summarize a list of simple and clear \textbf{design choices} for Diffuser{\footnotesize{Lite}}?

\vspace{-5pt}
\subsection{Experimental Setup} \label{sec:exp_setup}
\textbf{Benchmarks:} We evaluate the algorithm on various offline RL domains, including locomotion in Gym-MuJoCo \citep{greg2016gymmujoco}, real-world manipulation in FrankaKitchen \citep{gupta2020frankakitchen} and Robomimic \citep{robomimic}, long-horizon navigation in Antmaze \citep{fu2020d4rl}, and real-world stock trading in FinRL \citep{neorl}. We train all models using publicly available datasets (see \cref{append:domain} for further details).

\textbf{Baselines:} Our comparisons mainly include the basic imitation learning algorithm BC, existing offline RL methods CQL \citep{aviral2020cql} and IQL \citep{Ilya2022iql}, as well as two pioneering diffusion planning frameworks, Diffuser \citep{janner2022diffuser} and DD \citep{ajay2023dd}. Additionally, we compare with a state-of-the-art algorithm, HDMI \citep{li2023hdmi}, which is improved based on DD. (Further details about each baseline and the sources of performance results for each baseline across the experiments are presented in \cref{append:baselines}).

\textbf{Backbones:} As mentioned in Section \ref{sec:rf}, we implement three variations of Diffuser{\footnotesize Lite}, based on different backbones: 1) diffusion model, 2) rectified flow, and 3) rectified flow with an additional \textit{reflow} step. These variations will be used in subsequent experiments and indicated by the suffixes D, R1, and R2, respectively. All backbones utilize three levels, with a total temporal horizon of $129$ in MuJoCo and Antmaze, and $49$ in Kitchen. For more details about the hyperparameters selection for Diffuser{\footnotesize Lite}, please refer to \cref{append:implementation}.

\textbf{Computing Power:} All runtime results across our experiments are obtained on a server equipped with an Intel(R) Xeon(R) Gold 6326 CPU @ 2.90GHz and an NVIDIA GeForce RTX3090. \label{sec:compute_resource}

\vspace{-5pt}
\subsection{Runtime Cost~(RQ1)}

The primary objective of Diffuser{\footnotesize Lite} is to increase the decision-making frequency. Therefore, we first test the wall-clock runtime cost (time consumption for one action inference) of Diffuser{\footnotesize Lite} under three different backbones, compared to Diffuser, DD, and HDMI, to determine the extent of the advantage gained. We present the test results in \cref{tab:runtime} \footnote{Since the code for HDMI is not open-source, we make every effort to reproduce HDMI with the original settings to test its runtime cost and thus underline its results in \cref{tab:runtime}.}, which shows that the runtime cost of Diffuser{\footnotesize Lite} with D, R1, and R2 backbones is only $1.23\%$, $0.89\%$, and $0.51\%$ of the average runtime cost of Diffuser and DD, respectively. \textbf{The remarkable improvement in decision-making frequency does not harm its performance.} As shown in \cref{fig:4}, compared to the average success rates of Diffuser~(purple line) and DD~(grey line) on four FrankaKitchen sub-tasks,  Diffuser{\footnotesize Lite} improves them by $[41.5\%, 56.2\%, 55.3\%, 8.7\%]$, respectively, while being $200$-$300$ times faster. These improvements are attributed to ignoring redundant information in PRP, which reduces the complexity of the distribution that the backbone generative model needs to fit, allowing us to employ a light neural network backbone and use fewer sampling steps to conduct \textit{perfect-enough} planning. Its success in FrankaKitchen, a realistic robot manipulation scenario, also reflects its potential application in real-world settings.

\subsection{Performance~(RQ2)}

\begin{table*}[t]
\caption{\small{\textbf{D4RL Performance.} Results for Diffuser{\scriptsize Lite} correspond to the mean and standard error over $5$ random seeds. We detail the sources for the performance of prior methods in \cref{append:baselines}. Following Diffuser \citep{janner2022diffuser}, we emphasize in bold scores within $5$ percent of the maximum per task ($\ge 0.95\cdot\text{max}$).}}
\label{tab:performance_d4rl}
\vskip 0.0in
\begin{center}
\begin{small}
\scalebox{0.67}{
\begin{tabular}{llccccccccc}
\toprule
\textbf{Dataset} & \textbf{Environment} & \textbf{BC} & \textbf{CQL} & \textbf{IQL} & \textbf{Diffuser} & \textbf{DD} & \textbf{HDMI} & \textbf{Diffuser{\scriptsize Lite}-D} & \textbf{Diffuser{\scriptsize Lite}-R1} & \textbf{Diffuser{\scriptsize Lite}-R2} \\
\midrule

\multirow{3}{*}{Medium-Expert} 
& HalfCheetah & $55.2$  & $\bm{91.6}$  & $86.7$  & $79.8$  & $\bm{90.6}$  & $\bm{92.1}$  & $88.5\pm0.4$ & $\bm{90.8\pm0.9}$ & $84.0\pm2.9$ \\
& Hopper      & $52.5$  & $105.4$ & $91.5$  & $107.2$ & $\bm{111.8}$ & $\bm{113.5}$ & $\bm{111.6\pm0.2}$ & $\bm{110.3\pm0.3}$ & $\bm{110.1\pm0.5}$ \\
& Walker2d    & $\bm{107.5}$ & $\bm{108.8}$ & $\bm{109.6}$ & $\bm{108.4}$ & $\bm{108.8}$ & $\bm{107.9}$ & $\bm{107.1\pm0.6}$ & $\bm{106.4\pm0.3}$ & $\bm{106.1\pm0.7}$ \\
\midrule
\multirow{3}{*}{Medium} 
& HalfCheetah & $42.6$  & $44.0$  & $\bm{47.4}$  & $44.2$  & $\bm{49.1}$  & $\bm{48.0}$  & $\bm{48.9\pm1.1}$ & $\bm{48.6\pm0.7}$ & $45.3\pm0.5$ \\
& Hopper      & $52.9$  & $58.5$  & $66.3$  & $58.5$  & $79.3$  & $76.4$  & $\bm{100.9\pm1.1}$ & $\bm{99.5\pm0.7}$ & $\bm{96.8\pm0.3}$ \\
& Walker2d    & $75.3$  & $72.5$  & $78.3$  & $79.7$  & $82.5$  & $79.9$  & $\bm{88.8\pm0.6}$ & $\bm{85.1\pm0.5}$ & $83.7\pm1.0$ \\
\midrule
\multirow{3}{*}{Medium-Replay} 
& HalfCheetah & $36.6$  & $\bm{45.5}$  & $\bm{44.2}$  & $42.2$  & $39.3$  & $\bm{44.9}$  & $41.6\pm0.4$ & $42.9\pm0.4$ & $39.6\pm0.4$ \\
& Hopper      & $18.1$  & $95.0$  & $94.7$  & $\bm{96.8}$  & $\bm{100.0}$ & $\bm{99.6}$  & $\bm{96.6\pm0.3}$ & $\bm{97.8\pm1.3}$ & $93.2\pm0.7$ \\
& Walker2d    & $26.0$  & $77.2$  & $73.9$  & $61.2$  & $75.0$  & $80.7$  & $\bm{90.2\pm0.5}$ & $84.6\pm1.7$ & $78.2\pm1.7$ \\
\midrule
\multicolumn{2}{c}{\textbf{Average}}
& $51.9$ & $77.6$ & $77$ & $75.3$ & $\bm{81.8}$ & $\bm{82.6}$ & $\bm{86.0}$ & $\bm{85.1}$ & $\bm{81.9}$ \\
\midrule
Mixed   & Kitchen & $51.5$ & $52.4$ & $51.0$  & $50.0$ & $65$ & $69.2$ & $\bm{73.6\pm0.7}$ & $\bm{71.9\pm1.4}$ & $64.8\pm1.8$ \\
Partial & Kitchen & $38.0$ & $50.1$ & $46.3$  & $56.2$ & $57$ & $-$    & $\bm{74.4\pm0.6}$ & $69.9\pm0.7$ & $\bm{71.4\pm1.2}$ \\
\midrule
\multicolumn{2}{c}{\textbf{Average}}
& $44.8$ & $51.3$ & $48.7$ & $53.1$ & $61.0$ & $-$ & $\bm{74.0}$ & $\bm{70.9}$ & $68.1$ \\
\midrule
\multirow{2}{*}{Play} 
& Antmaze-Medium & $0.0$  & $65.8$  & $65.8$  & $0.0$  & $0.0$  & $-$  & $78.0\pm2.2$ & $\bm{88.0\pm2.2}$ & $\bm{88.8\pm3.2}$ \\
& Antmaze-Large  & $0.0$  & $20.8$  & $42.0$  & $0.0$  & $0.0$  & $-$  & $\bm{72.0\pm6.2}$ & $\bm{72.4\pm2.3}$ & $\bm{69.4\pm6.5}$ \\
\multirow{2}{*}{Diverse} 
& Antmaze-Medium & $0.8$  & $67.3$  & $73.8$  & $0.0$  & $0.0$  & $-$  & $\bm{92.4\pm3.2}$ & $\bm{89.2\pm2.0}$ & $\bm{87.6\pm2.0}$ \\
& Antmaze-Large  & $0.0$  & $20.5$  & $30.3$  & $0.0$  & $0.0$  & $-$  & $68.0\pm2.8$ & $\bm{80.4\pm5.1}$ & $75.2\pm3.5$ \\
\midrule
\multicolumn{2}{c}{\textbf{Average}}
& $0.2$ & $43.6$ & $53.0$ & $0.0$ & $0.0$ & $-$ & $77.6$ & $\bm{82.5}$ & $\bm{80.3}$ \\
\midrule
\multicolumn{2}{c}{\textbf{Runtime per action (second)}}
& $-$ & $-$ & $-$ & $0.749$ & $2.435$ & $-$ & $\bm{0.020}$ & $\bm{0.014}$ & $\bm{0.008}$ \\

\bottomrule
\end{tabular}}
\end{small}
\end{center}
\vspace{-10pt}
\end{table*}

\begin{table*}[t]
    \begin{minipage}[t]{0.48\linewidth}
        \caption{\small{\textbf{Robomimic Performance.}}}
        \label{tab:performance_robomimic}
        \centering
        \begin{small}
            \scalebox{0.64}{
                \begin{tabular}{lccccc}
                    \toprule
                    \textbf{Dataset} & \textbf{BC} & \textbf{CQL} & \textbf{BCQ} & \textbf{IRIS} & \textbf{Diffuser{\scriptsize Lite}-D} \\
                    \midrule
                    Lift-PH & $\bm{100.0}$ & $92.7$ & $\bm{100.0}$ & $\bm{100.0}$ & $\bm{100.0}$ \\
                    Can-PH &  $95.3$ & $38.0$ &  $88.7$ & $\bm{100.0}$ & $\bm{100.0}$ \\
                    Square-PH &  $78.7$ &  $5.3$ &  $50.0$ & $78.7$ &  $\bm{81.8}$ \\
                    \midrule
                    \textbf{Average} & $91.3$ & $45.3$ & $79.6$ & $92.9$ & $\bm{93.9}$  \\
                    \bottomrule
                \end{tabular}
            }
        \end{small}
    \end{minipage}
    \hfill
    \begin{minipage}[t]{0.48\linewidth}
        \caption{\small{\textbf{FinRL Performance.}}}
        \label{tab:performance_finrl}
        \centering
        \begin{small}
            \scalebox{0.64}{
                \begin{tabular}{lcccccc}
                    \toprule
                    \textbf{Dataset} & \textbf{BC} & \textbf{CQL} & \textbf{MB-PPO} & \textbf{DD} & \textbf{HDMI} & \textbf{Diffuser{\scriptsize Lite}-D} \\
                    \midrule
                    FinRL-H-999 & $270$ & $444$ & $787$ & $782$ & $\bm{801}$ & $796$ \\
                    FinRL-M-999 & $504$ & $621$ & $698$ & $712$ & $754$ & $\bm{762}$ \\
                    \midrule
                    \textbf{Average} & $387$ & $533$ & $743$ & $747$ & $778$ & $\bm{779}$ \\
                    \bottomrule
                \end{tabular}
            }
        \end{small}
    \end{minipage}
    \vspace{-10pt}
\end{table*}



 

Diffuser{\footnotesize Lite} is then evaluated on various popular domains in D4RL, Robomimic, and FinRL, to test how well it can maintain the performance when significantly increasing the decision-making frequency. All results are presented in \cref{tab:performance_d4rl}, \cref{tab:performance_robomimic}, and \cref{tab:performance_finrl}, and the detailed descriptions of the sources of all baseline results are listed in \cref{append:baselines}. Results in \textit{D4RL} table show significant performance improvements across all benchmarks with high decision-making frequency. This advantage is particularly pronounced in FrankaKitchen and Antmaze environments, indicating that the structure of Diffuser{\footnotesize Lite} enables more accurate and efficient planning in long-horizon tasks, thus yielding greater benefits. In the MuJoCo environments, more notable advantages are shown on sub-optimal datasets, i.e., ``medium" and ``medium-replay" datasets. This sub-optimal advantage can be attributed to the PRP planning structure, which does not require one-shot generation of a consistent long trajectory, but explicitly demands stitching. This allows for better utilization of high-quality segments in low-quality datasets, leading to improved performance. Results in \textit{Robomimic and FinRL} table are obtained by models trained on real-world datasets, and demonstrate that Diffuser{\footnotesize Lite} continues to exhibit its superiority in these real-world tasks, achieving performance comparable to SOTA algorithms. This illustrates the potential application of Diffuser{\footnotesize Lite} in real-world scenarios.

\vspace{-5pt}
\subsection{Flexible Plugin~(RQ3)}

\begin{table}
\caption{\small{\textbf{Integrate with Diffuser{\scriptsize Lite} as a plugin.} We refer to AlignDiff with Diffuser{\scriptsize Lite} plugin as AlignDiff-Lite. A larger value of \textit{MAE Area} indicates a stronger alignment capability. AlignDiff-Lite greatly increases decision-making frequency, while only experiencing a small performance drop.}}
\label{tab:aligndiff}
\vspace{-3pt}
\begin{center}
\begin{small}
\scalebox{0.77}{
\begin{tabular}{lccc}
\toprule
\textbf{Metric} & \textbf{GC} & \textbf{AlignDiff} & \textbf{AlignDiff-{\footnotesize Lite}}\\ 
\midrule
MAE Area & $0.319\pm0.005$ & $0.621\pm0.023$ & $0.601\pm0.018$ \textcolor{blue}{($3.2\%\downarrow$)} \\
Frequency~(Hz) & $-$ & $6.9$ & $45.5$ \textcolor{red}{($560\%\uparrow$)}\\
\bottomrule
\end{tabular}}
\end{small}
\end{center}
\vspace{-10pt}
\end{table}

\begin{table}[t]
    \centering
    \caption{\small{\textbf{Performance of Diffuser{\scriptsize Lite} with various PRP design choices.} The left part shows a comparison with 2/3/4 planning levels and the right part shows a comparison with 4 temporal horizon designs. Results correspond to the mean and standard error over 5 random seeds, the highest scores are emphasized in bold, and the default design choices used across other experiments are underlined.}}
    \label{tab:design_choice}
    \scalebox{0.77}{
    \begin{tabular}{lccc|cccc}
    \toprule
     & \multicolumn{7}{c}{\textbf{Temporal horizon of each level}} \\
    Planning horizon=$129$ & [9,17] & [5,5,9] & [5,3,5,5] & [3,5,17] & [5,5,9] & [9,5,5] & [17,5,3] \\
    \midrule
    HalfCheetah-me & $75.6\pm8.3$ & $\bm{\underline{88.5\pm0.4}}$ & $88.3\pm0.5$ & $85.6\pm0.6$ & $\underline{88.5\pm0.4}$ & $88.6\pm0.7$ & $\bm{89.0\pm1.7}$ \\
    Antmaze-ld     &  $0.0\pm0.0$ & $\underline{68.0\pm2.8}$ & $\bm{69.3\pm3.4}$ & $34.7\pm4.1$ & $\bm{\underline{68.0\pm2.8}}$ & $67.3\pm3.4$ & $34.0\pm4.3$ \\
    \midrule
    Planning horizon=$49$ & [7,9] & [4,5,5] & [3,4,3,5] & [3,3,13] & [4,5,5] & [5,5,4] & [13,3,3] \\
    \midrule
    Kitchen-p     &  $72.8\pm0.5$ & $\bm{\underline{74.4\pm0.6}}$ & $72.3\pm1.9$ & $66.7\pm1.7$ & $\bm{\underline{74.4\pm0.6}}$ & $74.2\pm0.6$ & $31.7\pm2.7$ \\
    \bottomrule
    \end{tabular}
    }
\vspace{-10pt}
\end{table}

To test the capability of Diffuser{\footnotesize Lite} as a flexible plugin to support other diffusion planning algorithms, Aligndiff \citep{dong2023aligndiff} is selected as a non-reward-maximizing algorithm backbone, and integrated with Diffuser{\footnotesize Lite} plugin, referred to as AlignDiff-Lite. AlignDiff aims to customize the agent's behavior to align with human preferences and introduces an MAE area metric to measure this alignment capability (refer to \cref{append:aligndiff} for more details), where a larger value indicates a stronger alignment capability. The comparison of AlignDiff and AlignDiff-Lite is presented in \cref{tab:aligndiff}, showing that AlignDiff-Lite achieves a $560\%$ improvement in decision-making frequency compared to AlignDiff, while only experiencing a small performance drop of $3.2\%$. This result demonstrates the potential of Diffuser{\footnotesize Lite} serving as a plugin to accelerate diffusion planning across various domains.

\vspace{-5pt}
\subsection{Ablations~(RQ4)}

\label{sec:design_choice}
\textbf{How to choose the number of planning levels $L$ and the temporal horizons $H_l$?} We compared the performance of Diffuser{\footnotesize Lite} with 2/3/4 planning levels and with four different temporal horizon designs, and reported the results in the left and right part of \cref{tab:design_choice}, respectively. The list in the first row represents the temporal horizon for each level, with larger values on the left indicating more planning at a coarser granularity, while larger values on the right indicate more planning at a finer granularity. For planning level design, results show a performance drop with 2 planning levels, particularly in Antmaze, and a consistently excellent performance with 3 or 4 planning levels. This suggests that for longer-horizon tasks, it is advisable to design more planning levels. For temporal horizon design, results show a performance drop when the temporal horizon of one level becomes excessively long. This suggests the temporal horizon of each level is supposed to be similar and stay close. We summarized and presented a design choice list in \cref{append:design_choice}.

\textbf{Has the last level (short horizon) of Diffuser{\footnotesize Lite} already performed well in decision making?} This is equivalent to the direct use of the shorter planning horizon. If it is true, key points generated by former levels may not have an impact, making PRP meaningless. To address this question, we conduct tests using a one-level Diffuser{\footnotesize Lite} with the same temporal horizon as the last level of the default model, referred to as \textbf{Lite} \textit{w/ only last level}. The results are presented in \cref{tab:PRP}, column $2$. The notable performance drop demonstrates the importance of a long-enough planning horizon.

\textbf{Can decision-making be effectively accomplished without progressive refinement planning (PRP)?} To address this question, we conduct tests using a one-level Diffuser{\footnotesize Lite} with the same temporal horizon as the default model, referred to as \textbf{Lite} \textit{w/o PRP}. This model only supports one-shot generation at the inference phase. We also test a ``smaller'' DD with the same network parameters and sampling steps as Diffuser{\footnotesize Lite}, to verify whether one can speed up DD by simply reducing the parameters, referred to as \textbf{DD}-\textit{small}. The results are presented in Table \ref{tab:PRP}, column $3$-$4$. The large standard deviation and the significant performance drop provide strong evidence for the limitations of one-shot generation planning, having difficulties in modeling the distribution of detailed long-horizon trajectories. However, Diffuser{\footnotesize Lite} can maintain high performance with fast decision-making frequency due to its lite architecture and simplified fitted distribution.

\textbf{More ablations.} We conducted further ablation studies on model size, sampling steps, planning horizon, with or without value condition, and visual comparison to better elucidate Diffuser{\footnotesize Lite}. Please refer to \cref{append:extra_exp} for these sections.

\begin{table}
\caption{\small{\textbf{Ablation tests conducted to examine the effectiveness of PRP.} \textbf{Lite} \textit{w/ only last level} and \textbf{Lite} \textit{w/o PRP} are two ablated versions of Diffuser{\scriptsize Lite}, while \textbf{DD}-\textit{small} is a version of DD that uses the same network parameters and sampling steps as Diffuser{\scriptsize Lite}. All results are obtained over $5$ seeds. The varying degrees of performance drop observed in each experiment highlight the importance of PRP.}}
\label{tab:PRP}
\vspace{-5pt}
\begin{center}
\begin{small}
\scalebox{0.71}{
\begin{tabular}{lcccc}
\toprule
\multirow{2}{*}{\textbf{Environment}} & \multirow{2}{*}{\textbf{Oracle}} & \textbf{Lite} \textit{w/} & \textbf{Lite} \textit{w/o} & \textbf{DD} \\
&& \textit{only last level}  & \textit{PRP} & \textit{small} \\
\midrule
Hopper-me & $111.6\pm0.2$ & $27.8\pm10.2$ & $96.6\pm1.0$ & $67.9\pm24.7$ \\
Hopper-m & $100.9\pm1.1$ & $19.4\pm2.0$ & $66.4\pm20.2$ & $16.7\pm8.0$ \\
Hopper-mr & $96.6\pm0.3$ & $2.0\pm0.2$ & $62.5\pm31.3$ & $1.0\pm0.4$ \\
\midrule
\textbf{Average} & $103.1$ & $16.4$ \textcolor{blue}{$(84.1\%\downarrow)$} & $75.2$ \textcolor{blue}{$(27.1\%\downarrow)$} & $28.5$ \textcolor{blue}{$(72.3\%\downarrow)$} \\
\midrule
Kitchen-m & $73.6\pm0.7$ & $48.2\pm1.4$ & $26.9\pm1.0$ & $0.0\pm0.0$ \\
Kitchen-p & $74.4\pm0.6$ & $38.6\pm3.1$ & $23.9\pm0.5$ & $1.8\pm0.8$ \\
\midrule
\textbf{Average} & $74.0$ & $43.4$ \textcolor{blue}{$(41.3\%\downarrow)$} & $25.4$ \textcolor{blue}{$(65.7\%\downarrow)$} & $0.8$ \textcolor{blue}{$(98.8\%\downarrow)$} \\
\bottomrule
\end{tabular}}
\end{small}
\end{center}
\vspace{-15pt}
\end{table}

\vspace{-8pt}
\section{Related Works}
\vspace{-8pt}

Diffusion models are a type of score-based generative model \citep{song2021scorebased}. Two pioneering frameworks, Diffuser \citep{janner2022diffuser} and DD \citep{ajay2023dd}, were the first to attempt using diffusion models for trajectory generation and planning in decision-making tasks. Based on these two frameworks, diffusion planning has been continuously improved and applied to various decision-making domains \citep{ni2023metadiffuser, du2023unipi, zhu2023madiff, dong2023aligndiff, he2023diffusion, lee2023RGG, jiang2023motiondiffuser}. However, long-horizon estimation and prediction often suffer from potential exponentially increasing variance concerning the temporal horizon, called the ``curse of horizon'' \citep{ren2021nearly}. To address this, HDMI \citep{li2023hdmi} is the first algorithm that proposed a hierarchical decision framework to generate sub-goals at the upper level and reach goals at the lower level, achieving improvements in long-horizon tasks. However, HDMI is limited by cluster-based dataset pre-processing to obtain high-quality sub-goal data for upper training. Another concurrent work, HD-DA \citep{chen2024hdda} introduces a similar hierarchical structure and allows the high-level diffusion model to automatically discover sub-goals from the dataset, achieving better results. However, the motivation behind Diffuser{\footnotesize Lite} is completely different, which aims to increase the decision-making frequency of diffusion planning. Also, Diffuser{\footnotesize Lite} allows for more hierarchy levels and refines only the first interval of the previous layer using PRP. Since HD-DA does not ignore redundant information, it fails to achieve a notable frequency increase. Compared to related works, Diffuser{\footnotesize Lite} has more clean plugin design and undoubtedly contributes to increasing decision-making frequency and performance. We believe it can serve as a reference for the design of future diffusion planning frameworks. 

\vspace{-5pt}
\section{Conclusion}

In this paper, we introduce Diffuser{\footnotesize Lite}, a super fast and lightweight diffusion planning framework that significantly increases decision-making frequency by employing the plan refinement process (PRP). PRP enables coarse-to-fine-grained trajectory generation, reducing the modeling of redundant information. Experimental results on various D4RL benchmarks demonstrate that Diffuser{\footnotesize Lite} achieves a super-high decision-making frequency of $122.2$Hz ($112.7$x faster than previous mainstream frameworks) while maintaining SOTA performance. Diffuser{\footnotesize Lite} provides three generative model backbones to adapt to different requirements and can be flexibly integrated into other diffusion planning algorithms as a plugin. \label{conclusion:limit}However, Diffuser{\footnotesize Lite} currently has limitations mainly caused by the classifier-free guidance (CFG). CFG sometimes requires adjusting the target condition, which becomes more cumbersome on the multi-level structure of Diffuser{\footnotesize Lite}. In future works, designing better guidance mechanisms, devising an optimal temporal jump adjustment, or integrating all levels in one diffusion model to simplify the framework is worth considering. Diffuser{\footnotesize Lite} may also have some societal impacts, such as expediting the deployment of robotic products that could be utilized for military purposes.

\newpage

\section{Acknowledgements}
This work is supported by the National Natural Science Foundation of China (Grant Nos. 62422605, 92370132).

\bibliography{neurips_2024}
\bibliographystyle{plainnat}


\newpage
\appendix

\section{Details of Experimental Setup}
\subsection{Test Domains}\label{append:domain}

\begin{figure}
    \centering
    \includegraphics[width=0.94\textwidth]{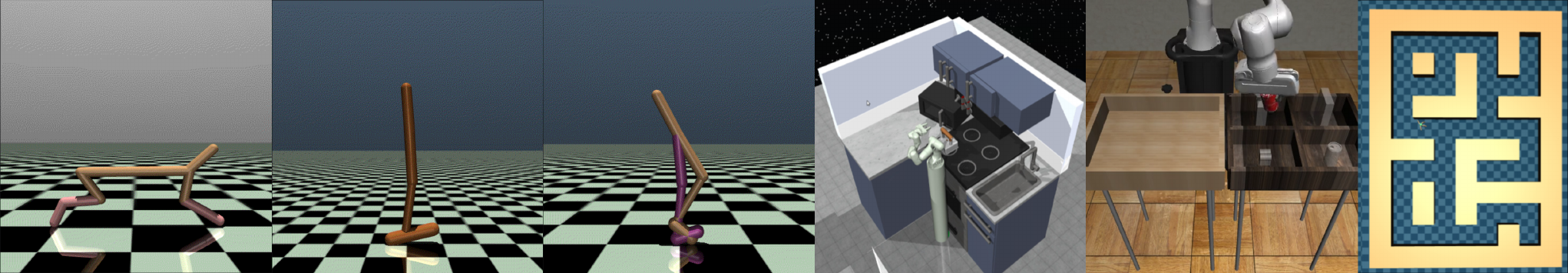}
    \caption{\small{Part of selected benchmarks. From left to right, they are HalfCheetah, Hopper, Walker2d, FrankaKitchen, Robomimic, and Antmaze.}}
    \label{fig:benchmarks}
\end{figure}

\textbf{Gym-MuJoCo} \citep{greg2016gymmujoco} on D4RL consists of three popular offline RL locomotion tasks (Hopper, HalfCheetah, Walker2d). These tasks require controlling three Mujoco robots to achieve maximum movement speed while minimizing energy consumption under stable conditions. D4RL provides three different quality levels of offline datasets: ``medium" containing demonstrations of ``medium" level performance, ``medium-replay" containing all recordings in the replay buffer observed during training until the policy reaches ``medium" performance, and ``medium-expert" which combines ``medium" and ``expert" level performance equally.

\textbf{FrankaKitchen} \citep{gupta2020frankakitchen} requires controlling a realistic 9-DoF Franka robot in a kitchen environment to complete several common household tasks. In offline RL testing, algorithms are often evaluated on ``partial" and ``mixed" datasets. The former contains demonstrations that partially solve all tasks and some that do not, while the latter contains no trajectories that completely solve the tasks. Therefore, these datasets place higher demands on the policy's ``stitching" ability. During testing, the robot's task pool includes four sub-tasks, and the evaluation score is based on the percentage of tasks completed.

\textbf{Antmaze} \citep{fu2020d4rl} requires controlling the $8$-DoF ``Ant" quadruped robot in MuJoCo to complete maze navigation tasks. In the offline dataset, the robot only receives a reward upon reaching the endpoint, and the dataset contains many trajectory segments that do not lead to the endpoint, making it a difficult decision task with sparse rewards and a long horizon. The success rate of reaching the endpoint is used as the evaluation score, and common model-free offline RL algorithms often struggle to achieve good performance.

\textbf{Robomimic} \citep{robomimic} requires learning policies to complete complex manipulation tasks from a small number of human demonstrations. Due to the non-Markovian nature of human demonstrations and the demonstration quality variance, learning from human datasets is significantly more challenging than learning from machine-generated datasets. In our experiments, we used a dataset of PH (Proficient-Human) type, which consists of 200 demonstrations collected through teleoperation by one single experienced teleoperator.

\textbf{FinRL} \citep{neorl} provides a way to build a trading simulator that replicates the real stock market and supports backtesting with important market frictions such as transaction costs, market liquidity, and investor risk aversion, among other factors. In the FinRL environment, one trade can be made per trading day for the stocks in the pool (30 stocks). The reward function is the difference in the total asset value between the end of the day and the day before. The environment may evolve itself as time elapses. In our experiments, the dataset has two suffixes, where ``H'' and ``M'' respectively indicate the quality of the dataset (``High'' and ``Medium''), and 999 indicates that the dataset includes 999 rollouts.

\subsection{Baselines}\label{append:baselines}

\subsubsection{Runtime Testing}
\begin{itemize}
    \item We run Diffuser \footnote{\href{https://github.com/jannerm/diffuser}{https://github.com/jannerm/diffuser}} using the official repository from the original paper with default hyperparameters.
    \item We run DD \footnote{\href{https://github.com/anuragajay/decision-diffuser/tree/main/code}{https://github.com/anuragajay/decision-diffuser/tree/main/code}} using the official repository from the original paper with default hyperparameters.
    \item Since the code for HDMI has not been released, we made every effort to reproduce HDMI based on the Diffuser{\footnotesize Lite} codebase, strictly adhering to every detail mentioned in the paper.
\end{itemize}
All runtime results are obtained on a server equipped with an Intel(R) Xeon(R) Gold 6326 CPU @ 2.90GHz and an NVIDIA GeForce RTX3090.

\subsubsection{Reward-maximizing}
\begin{itemize}
    \item The performance of BC, CQL \citep{aviral2020cql} and IQL \citep{Ilya2022iql} in \cref{tab:performance_d4rl} is reported in the D4RL \citep{fu2020d4rl}, Table 2;
    \item The performance of Diffuser \citep{janner2022diffuser} in \cref{tab:performance_d4rl}, MuJoCo, is reported in \citep{janner2022diffuser}, Table 1; The performance in Kitchen is reported in \citep{chen2024hdda}, Table 2; And the performance in Antmaze is obtained by the official repository from the original paper with default hyperparameters.
    \item The performance of DD \citep{ajay2023dd} in \cref{tab:performance_d4rl}, MuJoCo and Kitchen, is reported in \citep{ajay2023dd}, Table 1; And the performance in Antmaze is reported in \citep{mineui2023dtamp}, Table 1;
    \item The performance of HDMI \citep{li2023hdmi} in \cref{tab:performance_d4rl} is reported in \citep{li2023hdmi}, Table 3.
\end{itemize}

\subsubsection{Behavior-customizing}
\begin{itemize}
    \item The performance of GC (goal conditioned BC) \citep{emmons2022rvs} and AlignDiff \citep{dong2023aligndiff} in \cref{tab:aligndiff} is reported in \citep{dong2023aligndiff}, Table 2.
\end{itemize}

\subsection{Details of AlignDiff}\label{append:aligndiff}

\textbf{Why AlignDiff?} Since Diffuser{\footnotesize Lite} can be seen as integrating PRP into DD, the main experiments in the paper can be considered as results obtained when Diffuser{\footnotesize Lite} is used as a reward-maximizing algorithm plugin. Therefore, to evaluate the more general plugin capability of Diffuser{\footnotesize Lite}, a non-reward-maximizing algorithm backbone must be selected. AlignDiff, unlike the common setting in Offline RL, uses diffusion planning for behavior-customizing, which aligns with our requirements.

AlignDiff \citep{dong2023aligndiff} is a diffusion planning algorithm used for behavior-customizing. In our experiments, we integrate Diffuser{\footnotesize Lite} as a plugin into AlignDiff, achieving a significant increase in decision-making frequency of approximately $560\%$ with minimal performance loss (around $3.2\%$). This demonstrates the flexibility and effectiveness of Diffuser{\footnotesize Lite} serving as a plugin. In this section, we provide a detailed description of the problem setting for AlignDiff and the evaluation metric used to assess the algorithm's performance, aiming to help the understanding of the experimental content.

\textbf{Problem setting:} AlignDiff considers a reward-free Markov Decision Process (MDP) denoted as $\mathcal{M} = \langle S, A, P, \bm{\alpha} \rangle$. Here, $S$ represents the set of states, $A$ represents the set of actions, $P: S \times A \times S \rightarrow [0, 1]$ is the transition function, and $\bm{\alpha} = \{\alpha_1, \cdots, \alpha_k\}$ represents a set of $k$ predefined attributes used to characterize the agent's behaviors. Given a state-only trajectory $\tau^l = \{s_0, \cdots, s_{l-1}\}$, it assumes the existence of an attribute strength function that maps the trajectory to a relative strength vector $\zeta^{\bm{\alpha}}(\tau^l) = \bm{v}^{\bm{\alpha}} = [v^{\alpha_1}, \cdots, v^{\alpha_k}] \in [0, 1]^k$. Each element of the vector indicates the relative strength of the corresponding attribute. A value of $0$ for $v^{\alpha_i}$ implies the weakest manifestation of attribute $\alpha_i$, while a value of $1$ represents the strongest manifestation. Human preferences are formulated as a pair of vectors $(\bm{v}_{\text{targ}}^{\bm{\alpha}}, \bm{m}^{\bm{\alpha}})$, where $\bm{v}_{\text{targ}}^{\bm{\alpha}}$ represents the target relative strengths, and $\bm{m}^{\bm{\alpha}} \in \{0, 1\}^k$ is a binary mask indicating which attributes are of interest. The objective is to find a policy $a=\pi(s|\bm{v}_{\text{targ}}^{\bm{\alpha}}, \bm{m}^{\bm{\alpha}})$ that minimizes the L1 norm $||(\bm{v}_{\text{targ}}^{\bm{\alpha}} - \zeta^{\bm{\alpha}}(\mathbb{E}_{\pi}[\tau^l])) \circ \bm{m}^{\bm{\alpha}}||_1$, where $\circ$ denotes the Hadamard product.

\textbf{Area metric:} To evaluate the algorithm's performance, the authors suggest that one can conduct multiple trials to collect the mean absolute error (MAE) between the evaluated and target relative strengths. For each trial, we need to sample an initial state $s_0$, a target strengths $\bm v^{\bm \alpha}_{\text{targ}}$, and a mask $\bm m^{\bm\alpha}$, as conditions for the execution of each algorithm. Subsequently, the algorithm runs for $T$ steps, resulting in the exhibited relative strengths $\bm v^{\bm \alpha}$ evaluated by $\hat\zeta_\theta$. Then we can calculate the percentage of samples that fell below pre-designed thresholds to create an MAE curve. The area enclosed by the curve and the axes can be used to define an area metric. A larger metric value indicates better performance in matching. By integrating Diffuser{\footnotesize Lite}, AlignDiff can maintain almost the same level of performance with only a $3.2\%$ decrease. The overall performance is nearly twice as good as the goal-conditioned behavior clone (GC) while achieving a significant increase in decision frequency of $560\%$.

\section{Details of Rectified Flow}\label{append:rectifiedflow}
Similar to DMs, rectified flow \citep{liu2023rectifiedflow} is also a probability flow-based generative model that learns a transfer from $q_0$ to $q_1$ through an ODE. In trajectory generation, we can define $q_1$ as the distribution of trajectories and $q_0$ as the standard Gaussian. The learned ODE can be represented as:
\begin{equation}
    \dd\bm x_s = v(\bm x_s, s) ds, \text{ initialized from }\bm x_0\sim q_0, \text{ such that }\bm x_1\sim q_1
\end{equation}

where $v: \mathbb R^d\times[0,1]\rightarrow\mathbb R^d$ is a velocity field, learned by minimizing a simple mean square objective:
\begin{equation}\label{eq:rf_training}
    \min_v\mathbb E_{(\bm x_0,\bm x_1)\sim\gamma}[\int_0^1||\frac{\dd}{\dd s}\bm x_s-v(\bm x_s,s)||^2\dd s], \text{ with } \bm x_s=(1-s)\bm x_0+s\bm x_1
\end{equation}
where $\gamma$ is any coupling of $(q_0,q_1)$, and $v$ is parameterized as a deep neural network and \cref{eq:rf_training} is solved approximately with stochastic gradient methods. A key property of rectified flow is its ability to learn a straight flow, which means:
\begin{equation}
    \textit{Straight flow: } \bm x_s=s\bm x_s+(1-s)\bm x_0 = \bm x_0+sv(\bm x_0,0), \forall s\in[0,1]
\end{equation}
A straight flow can achieve \textit{perfect} results with fewer sample steps (even a single step).

\textit{Reflow} is an iterative procedure to straighten the learned flow without modifying the marginal distributions, hence allowing faster sampling at inference time. Assume we have an ODE model $\dd\bm x_s=v_k(\bm x_s,s)\dd s$ with velocity field $v_k$ at the $k$-th iteration of the reflow procedure; denote by $\bm x_1=\texttt{ODE}[v_k](\bm x_0)$ the $\bm x_s$ we obtained at $t=1$ when following the $v_k$-ODE starting from $\bm 0$. A reflow step turns $v_k$ into a new vector field $v_{k+1}$ that yields straighter ODEs while $\bm x_1^{\text{new}}=\texttt{ODE}[v_{k+1}](\bm x_0)$ has the same distribution as $\bm x_1=\texttt{ODE}[v_{k}](\bm x_0)$,
\begin{multline}
    v_{k+1}=\mathop{\arg\min}\limits_{v} \mathbb E_{\bm x_0\sim{q_0}}[\int_0^1||(\bm x_1-\bm x_0)-v(\bm x_s,s)||^2\dd s], \\
\text{ with }\bm x_1=\texttt{ODE}[v_k](\bm x_0)\text{ and }\bm x_s=s\cdot\bm x_1 + (1-s)\cdot\bm x_0,
\end{multline}
where $v_{k+1}$ is learned using the same rectified flow objective \cref{eq:rf_training}, but with the linear interpolation of $(\bm x_0, \bm x_1)$ pairs constructed from the previous $\texttt{ODE}[v_k]$. For conditional sampling, rectified flow also supports classifier-free guidance, in which we can train a conditional velocity field $v(\bm x_s, s | \bm c)$ and apply CFG by:
\begin{equation}
    v^w(\bm x_s, s|\bm c)=wv(\bm x_s,s|\bm c)+(1-w)v(\bm x_s, s),
\end{equation}
where $w$ is the guidance strength. In this way, we can directly replace diffusion models with rectified flow as the backbone of Diffuser{\footnotesize Lite}. In our experiments, we find that rectified flow achieves similar performance to the diffusion backbone when using the same number of sampling steps and neural network size. By further conducting a \textit{reflow} procedure, the planning of the model becomes more stable under the same number of sampling steps (as evidenced by a decrease in the variance of experimental results). It is even possible to reduce the number of sampling steps to just one, resulting in only a small performance drop.

\section{PRP Design Insights}\label{append:design_choice}

As the core planning framework of Diffuser{\footnotesize Lite}, PRP design choices are crucial. Through experimental results and discussions in \cref{sec:design_choice}, some simple, stable, and effective PRP design insights can be summarized:
\begin{itemize}
    \item The planning horizon $T$ is supposed to be determined by the nature of the task. The sparser the rewards and the longer the episodes, the longer the planning horizon should be used.
    \item The number of planning levels $L$ is supposed to be designed based on the planning horizon, with longer horizon tasks requiring more planning levels. One can start from 3 or 4 planning levels.
    \item The temporal horizon of each level $H_l$ is supposed to be similar and limited to within 10 to increase decision-making frequency. 
    \item Once the number of planning levels and temporal horizon are determined, the temporal jump is also determined.
\end{itemize}
Throughout our experiments, we consistently applied this set of design choices, achieving consistently excellent performance without the need for carefully adjusting parameters.

\section{Implementation Details}\label{append:implementation}
We introduce the implementation details of Diffuser{\footnotesize Lite} in the section:
\begin{itemize}
    \item We utilize DiT \citep{Peebles2022DiT} as the neural network backbone for all diffusion models and rectified flows, with an embedding dimension of $256$, $8$ attention heads, and $2$ DiT blocks. We progressively reduce the model size from DiT-S \citep{Peebles2022DiT} until the current size setting, and there is still no significant performance drop. This suggests that future works could even explore further reduction of network parameters to achieve faster decision speeds.
    \item Across all the experiments, we employ Diffuser{\footnotesize Lite} with $3$ levels. In Kitchen, we utilize a planning horizon of $49$ with temporal jumps for each level set to $16$, $4$, and $1$, respectively. In MuJoCo and Antmaze, we use a planning horizon of $129$ with temporal jumps of $32$, $8$, and $1$ for each level, respectively.
    \item For diffusion models, we use cosine noise schedule \citep{nichol2021cosineschedule} for $\alpha_s$ and $\sigma_s$ with diffusion steps $T=1000$. We employ DDIM \citep{song2021DDIM} to sample trajectories. In MuJoCo and Kitchen, we use $3$ sampling steps, while in Antmaze, we use $5$ sampling steps.
    \item For rectified flows, we use the Euler solver with $3$ steps for all benchmarks. After one reflow procedure, we can further reduce it to $1$ step for MuJoCo and $2$ steps for Kitchen and Antmaze.
    \item All models utilize the AdamW optimizer \citep{ilya2019adamw} with a learning rate of $2e-4$ and weight decay of $1e-5$. We perform $500$K gradient updates with a batch size of $256$. We do not employ the exponential moving average (EMA) model, as used in \citep{janner2022diffuser,ajay2023dd}. We found that using the EMA model did not yield significant gains in our experiments. For \textit{reflow} training, we first use the trained rectified flow to generate a $2$M dataset with $20$ sampling steps. Then we use the same optimizer, but with a learning rate of $2e-5$, to train the model for $200$K gradient steps.
    
    \item For conditional sampling, we tune the guidance strength $w$ within the range of $[0,1]$. In general, a higher guidance strength leads to better performance but may result in unrealistic plans and instability. On the other hand, a lower guidance strength provides more stability but may lead to a decrease in performance. \textbf{In our implementation of Diffuser{\footnotesize Lite}, we observe that the earlier levels are more closely related to decision-making. In contrast, the later levels only need to ensure reaching the key points provided by the previous levels. Therefore, we only apply conditional sampling to level $\bm 0$, while the other levels are not guided.}
    \item In MuJoCo, we only utilize the cumulative rewards as the generation condition. However, in Kitchen and Antmaze, we employ the discounted cumulative return and the value of the last generated state as the generation condition. The values are evaluated using a pre-trained IQL-value function \citep{Ilya2022iql}. We have found that this generation condition is beneficial for training models in reward-sparse environments, as it helps to prevent the model from becoming confused by suboptimal or poor trajectories.
    \item The inverse dynamic model is implemented as a 3-layer MLP. The first two layers consist of a Linear layer followed by a Mish activation \citep{diganta2020Mish} and a LayerNorm \citep{lei2016layernorm}. And, the final layer is followed by a Tanh activation. This model utilizes the same optimizer as the diffusion models and is trained for $200$K gradient steps.
\end{itemize}

\section{Additional Experiment Results}\label{append:extra_exp}

\subsection{Performance with Different Model Sizes}

We compare the performance of Diffuser{\footnotesize Lite} using four different model sizes. Due to the use of DiT as the backbone, we fix the dimensions of each attention head, and the parameter design of the Transformer is as follows: [\texttt{hidden\_size}, \texttt{n\_heads}, \texttt{depth}] respectively set to [128, 4, 2] (0.68M), [192, 6, 2] (1.53M), [256, 8, 2] (2.7M), and [320, 10, 2] (4.22M). The results are presented in \cref{tab:model_sizes}, which shows that slightly increasing the model size does not significantly decrease the inference speed, but it does improve the performance. This suggests that increasing the model parameter size can further enhance the performance of Diffuser{\footnotesize Lite}.

\begin{table}
\caption{\small{\textbf{Performance of Diffuser{\scriptsize Lite} under four different model sizes.} All results are obtained over $5$ seeds. The results under default choice are underlined and the highest scores are emphasized in bold.}}
\label{tab:model_sizes}
\vskip 0.0in
\begin{center}
\begin{small}
\scalebox{0.9}{
\begin{tabular}{lcccc}
\toprule
\multirow{2}{*}{\textbf{Environment}} & \multicolumn{4}{c}{\textbf{Model sizes}} \\
\cmidrule{2-5}
 & \textbf{0.68M} & \textbf{1.53M} & \underline{\textbf{2.7M}} & \textbf{4.22M}\\ 
\midrule
HalfCheetah-me & $75.4\pm5.1$ & $86.9\pm0.6$  & $\underline{88.5\pm0.4}$ & $\bm{90.6\pm1.1}$ \\
Runtime (s) & $\bm{0.0153}$ & $0.0154$ & $\underline{0.0155}$ & $0.0156$ \\
\midrule
Antmaze-ld & $0.3\pm0.5$ & $60.7\pm2.5$  & $\underline{68.0\pm2.8}$ & $\bm{73.0\pm3.6}$ \\
Runtime (s) & $\bm{0.0262}$ & $0.0268$ & $\underline{0.0270}$ & $0.0271$ \\
\midrule
Kitchen-p & $64.5\pm0.8$ & $74.8\pm0.2$  & $\underline{74.4\pm0.6}$ & $\bm{75.0\pm0.0}$ \\
Runtime (s) & $\bm{0.0163}$ & $0.0169$ & $\underline{0.0170}$ & $0.0172$ \\

\bottomrule
\end{tabular}}
\end{small}
\end{center}
\vskip -0.1in
\end{table}

\subsection{Performance with Different Sampling Steps}

We compare the performance of Diffuser{\footnotesize Lite} using four different sampling steps. The results are presented in \cref{tab:sample_steps}, which shows a natural trade-off: more sampling steps lead to better performance but at the cost of slower decision-making speed. Diffuser{\footnotesize Lite} strikes a balance between performance and speed by offering the choice of 3 or 5 sampling steps. Researchers can flexibly adjust the sampling steps based on the specific requirements of their tasks.

\begin{table}
\caption{\small{\textbf{Performance of Diffuser{\scriptsize Lite} with different sampling steps.} All results are obtained over $5$ seeds. The results under default choice are underlined and the highest scores are emphasized in bold.}}
\label{tab:sample_steps}
\vskip 0.0in
\begin{center}
\begin{small}
\scalebox{0.9}{
\begin{tabular}{lcccc}
\toprule
\multirow{2}{*}{\textbf{Environment}} & \multicolumn{4}{c}{\textbf{Sampling steps}} \\
\cmidrule{2-5}
 & \textbf{1} & \underline{\textbf{3}} & \underline{\textbf{5}} & \textbf{10}\\ 
\midrule
HalfCheetah-me & $74.4\pm15.7$ & $\underline{88.5\pm0.4}$  & ${89.0\pm0.7}$ & $\bm{89.2\pm1.3}$ \\
Runtime (s) & $\bm{0.0065}$ & $\underline{0.0155}$ & ${0.026}$ & $0.048$ \\
\midrule
Antmaze-ld & $0.7\pm0.9$ & $16.7\pm2.5$  & $\underline{68.0\pm2.8}$ & $\bm{74.3\pm5.4}$ \\
Runtime (s) & $\bm{0.0068}$ & $0.0167$ & $\underline{0.0270}$ & $0.050$ \\
\midrule
Kitchen-p & $5.7\pm1.2$ & $\underline{74.4\pm0.6}$  & ${73.8\pm0.8}$ & $\bm{74.7\pm0.5}$ \\
Runtime (s) & $\bm{0.0068}$ & $\underline{0.0170}$ & ${0.029}$ & $0.053$ \\

\bottomrule
\end{tabular}}
\end{small}
\end{center}
\vskip -0.1in
\end{table}

\subsection{Performance under Different Planning Horizon Choices}

The planning horizon is an essential parameter that influences the performance of planning algorithms. To investigate the performances of Diffuser{\footnotesize Lite} under different temporal horizon choices, we test it using three temporal horizon choices: $49$, $129$, and $257$. The results are presented in \cref{tab:horizons}. In Hopper environment, a longer planning horizon is required to avoid greedy and rapid jumps that may lead to falls. Consequently, the performance of Diffuser{\footnotesize Lite} is slightly poorer under the $49$ temporal horizon compared to the other two choices ($129$ and $257$), where no significant performance differences are observed. For Kitchen environment, the total length of the episode ($280$ time steps) in the dataset poses a limitation. An excessively long planning horizon can confuse the model, as it may struggle to determine the appropriate actions to take after completing all tasks. As a result, the performance of Diffuser{\footnotesize Lite} is poor under the $257$ temporal horizon, while no significant performance differences are observed under the $49$ and $129$ temporal horizons.

\subsection{Performance with or without Value Condition}

\begin{figure}[H]
    \centering
    \includegraphics[width=0.5\textwidth]{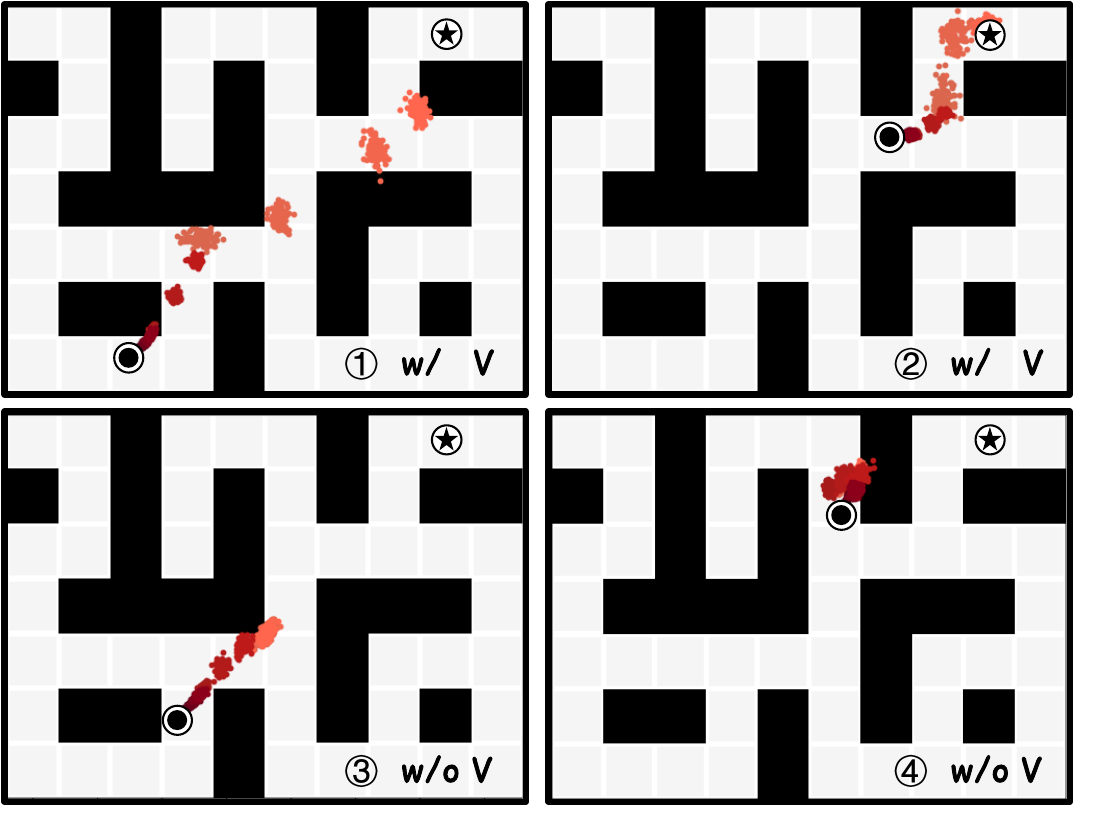}
    \vspace{-4pt}
    \caption{\small{\textbf{Visual comparison of Diffuser{\scriptsize Lite} using conditions with (upper) or without (lower) values.} It displays $100$ plans generated at the current state, where darker colors indicate closer proximity to the current state, and lighter colors indicate further. With the pure-rewards condition, we can observe that the planned states in lighter colors tend to cluster together at a certain point on the map, indicating the planner tries to stop at that non-endpoint. However, with the introduction of values, the planner can make correct long-term plans that lead to the endpoint.}}
    \label{fig:wwoV}
    \vspace{-10pt}
\end{figure}

Diffuser{\footnotesize Lite} offers two optional generation conditions, resulting in two different critics, as introduced in \cref{sec:4}. To determine the suitability of these two approaches in different tasks, we conduct ablation experiments and present the results in \cref{tab:woV}. We find that in dense reward tasks, such as Hopper and HalfCheetah, the performance of both conditions is nearly identical. However, in sparse reward scenarios, such as Kitchen and Antmaze, the use of values demonstrates a significant advantage. We present a visual comparison in Antmaze in \cref{fig:wwoV}, which shows $100$ generated plans. With a pure-rewards condition, it is difficult to discern the endpoint location, as the planner often desires to stop at a certain point on the map. However, when using values, the planner indicates a desire to move towards the endpoint. This suggests that sparse reward tasks are prone to confusing the conditional generative model, leading to poor planning. The introduction of value-assisted guidance can address this issue.

\subsection{Visual Comparison between One-shot Generated Plans and PRP}
\begin{figure}
    \centering
    \includegraphics[width=0.94\textwidth]{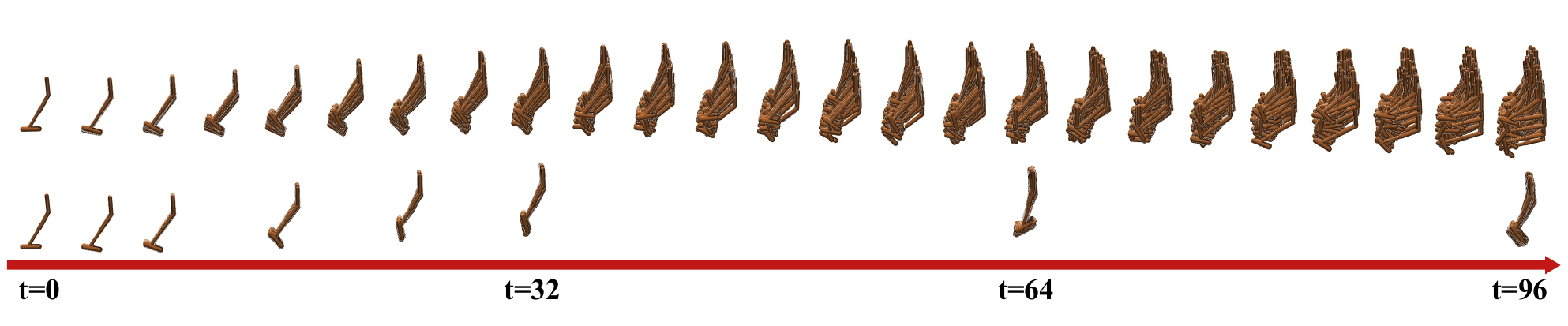}\label{fig:hopper}
    \caption{\small{\textbf{Visual comparison between one-shot generated plans (upper) and PRP (lower) in Hopper.} The figure showcases $100$ diffusion-generated plans starting from the same initial state. The one-shot generated plans directly generate all states from $t=0$ to $t=96$, while PRP utilizes three levels with temporal horizons of $97$, $33$, and $9$, and temporal jumps of $32$, $8$, and $1$, respectively. For ease of observation, only Hopper motion every $4$ steps is displayed in the figure.}}
\end{figure}

We visually compare the one-shot generation and PRP in Hopper, presenting the results in \cref{fig:hopper}. Regarding efficiency, PRP significantly reduces the length of sequences that need to be generated due to each level refining only the first jumpy interval of the previous level. This results in a noticeable increase in the forward speed of neural networks, especially for transformer-based backbone models like DiT \citep{Peebles2022DiT}. Visually, we can more clearly perceive the difference in generated sequence lengths, and this advantage will further expand as the total planning horizon increases. In terms of quality, PRP narrows down the search space of planning, leading to better consistency among different plans, while the one-shot generated plans exhibit more significant divergence in the far horizon.

\begin{table}
\caption{\small{\textbf{Performance of Diffuser{\scriptsize Lite} under three different temporal horizon choices.} All results are obtained over $5$ seeds and the results of default choice are emphasized in bold scores.}}
\label{tab:horizons}
\vskip 0.0in
\begin{center}
\begin{small}
\scalebox{0.9}{
\begin{tabular}{lccc}
\toprule
\multirow{2}{*}{\textbf{Environment}} & \multicolumn{3}{c}{\textbf{Temporal Horizon - 1}} \\
\cmidrule{2-4}
 & \textbf{48} & \textbf{128} & \textbf{256}\\ 
\midrule
Hopper-me & $101.1\pm0.5$ & $\bm{111.6\pm0.2}$  & $110.0\pm0.4$\\
Hopper-m  & $96.6\pm10.9$ & $\bm{100.9\pm1.1}$ & $98.9\pm0.4$ \\
Hopper-mr & $74.5\pm27.2$ & $\bm{96.6\pm0.3}$ & $98.2\pm2.1$ \\
\midrule
\textbf{Average} & $90.7$ & $\bm{103.1}$ & $102.4$ \\
\midrule
Kitchen-m & $\bm{73.6\pm0.7}$ & $72.7\pm1.0$ & $25.1\pm0.2$ \\
Kitchen-p & $\bm{74.4\pm0.6}$ & $75.0\pm0.0$ & $25.2\pm0.4$ \\
\midrule
\textbf{Average} & $\bm{74.0}$ & $73.9$ & $25.2$ \\
\bottomrule
\end{tabular}}
\end{small}
\end{center}
\vskip -0.1in
\end{table}

\begin{algorithm}[t]
   \caption{Diffuser{\footnotesize Lite} Training}
   \label{alg:training}
\begin{algorithmic}
   \STATE {\bfseries Input:} number of planning levels $L$, temporal horizon $H_l$, temporal jump $I_l$ and noise estimators $\bm\epsilon_{\theta_l}$ for each level $l\in\{0,1,\cdots,L-1\}$, dataset $\mathcal{D}=\{\bm x\}$, where $\bm x$ is the sequence of state-action pairs, critic $\mathcal{C}$, diffusion steps $T$, condition mask probability $1-p$;
   \WHILE {not done}
   \STATE sample a batch of $(\bm x, \mathcal{C}(\bm x))$ from dataset $\mathcal{D}$
    \FOR{$l=0$ {\bfseries to} $L-1$}
    \STATE extract $\bm x_{0,H_l,1}$ and $\bm x_{0,H_l,I_l}$ from $\bm x$
    \STATE $\hat{\mathcal{C}}({\bm x_{0,H_l,I_l})}\leftarrow \mathcal{C}(\bm x_{0,H_l,1})$
    \STATE sample $s_l\sim\text{Uniform}(T)$, $\bm\epsilon_l\sim \mathcal{N}(\bm 0, \bm I)$
    \STATE first state of $\bm\epsilon_l\leftarrow$ first state of $\bm x_{0,H_l,I_l}$
    \IF{$l>0$}
    \STATE last state of $\bm\epsilon_l\leftarrow$ last state of $\bm x_{0,H_l,I_l}$
    \ENDIF
    \STATE $\bm x^s_{0:H_l:I_l}\leftarrow\alpha_s\bm x_{0:H_l:I_l}+\sigma_s\bm\epsilon_l$
    \STATE update $\theta_l$ by minimizing \\ $||\bm\epsilon_{\theta_l}(\bm x^s_{0:H_l:I_l},s,\mathcal{C}(\bm x_{0:H_l:I_l}))-\bm\epsilon||^2$ with probability $p$ else \\$||\bm\epsilon_{\theta_l}(\bm x^s_{0:H_l:I_l},s))-\bm\epsilon||^2$ 
    \ENDFOR
    \ENDWHILE
\end{algorithmic}
\end{algorithm}

\begin{algorithm}[t]
   \caption{Diffuser{\footnotesize Lite} Inference}
   \label{alg:inference}
\begin{algorithmic}
   \STATE {\bfseries Input:} number of planning levels $L$, temporal horizon $H_l$, temporal jump $I_l$ and noise estimators $\bm\epsilon_{\theta_l}$ for each level $l\in\{0,1,\cdots,L-1\}$, critic $\mathcal{C}$, inverse dynamic $h$, diffusion steps $T$, current state $o_t$;
   \FOR{$l=0$ {\bfseries to} $L-1$}
   \STATE sample $\bm\epsilon_l\sim\mathcal{N}(\bm 0,\bm I)$.
   \STATE first state of $\bm\epsilon_l\leftarrow o_t$ 
   \IF{$l>0$}
   \STATE last state of $\bm\epsilon_l\leftarrow$ last state of $\bm x_{0:H_{l-1}:I_{l-1}}$
   \ENDIF
   \STATE obtain $\bm x_{0:H_l:I_l}$ by solving diffusion ODE with DDIM solver
   \ENDFOR
   \STATE extract $o_t, o_{t+1}$ from $\bm x_{0:H_{L-1}:1}$
   \STATE $a_t = h(o_t, o_{t+1})$
\end{algorithmic}
\end{algorithm}

\begin{table}[H]
\caption{\small{\textbf{Performance of Diffuser{\scriptsize Lite} using conditions with or without values.} Present only average performance on varying-quality datasets, which are obtained over $5$ seeds.}}
\label{tab:woV}
\vskip 0.0in
\begin{center}
\begin{small}
\scalebox{0.9}{
\begin{tabular}{lcccc}
\toprule
\textbf{Condition} & \textbf{Hopper} & \textbf{HalfCheetah} & \textbf{Kitchen} & \textbf{Antmaze} \\
\midrule
w/~~ Value & $103.6$ & $60.7$ & $74.0$ & $77.6$ \\
w/o Value & $103.1$ & $59.7$ & $54.1$ & $19.7$ \\
\bottomrule
\end{tabular}}
\end{small}
\end{center}
\vskip -0.1in
\end{table}

\newpage

\section*{NeurIPS Paper Checklist}

\begin{enumerate}

\item {\bf Claims}
    \item[] Question: Do the main claims made in the abstract and introduction accurately reflect the paper's contributions and scope?
    \item[] Answer: \answerYes{}
    \item[] Justification: See the abstract's last three sentences and the introduction's last paragraph in \Cref{intro:claims}.
    \item[] Guidelines:
    \begin{itemize}
        \item The answer NA means that the abstract and introduction do not include the claims made in the paper.
        \item The abstract and/or introduction should clearly state the claims made, including the contributions made in the paper and important assumptions and limitations. A No or NA answer to this question will not be perceived well by the reviewers. 
        \item The claims made should match theoretical and experimental results, and reflect how much the results can be expected to generalize to other settings. 
        \item It is fine to include aspirational goals as motivation as long as it is clear that these goals are not attained by the paper. 
    \end{itemize}

\item {\bf Limitations}
    \item[] Question: Does the paper discuss the limitations of the work performed by the authors?
    \item[] Answer: \answerYes{}
    \item[] Justification: See the conclusion in \Cref{conclusion:limit}.
    \item[] Guidelines:
    \begin{itemize}
        \item The answer NA means that the paper has no limitation while the answer No means that the paper has limitations, but those are not discussed in the paper. 
        \item The authors are encouraged to create a separate "Limitations" section in their paper.
        \item The paper should point out any strong assumptions and how robust the results are to violations of these assumptions (e.g., independence assumptions, noiseless settings, model well-specification, asymptotic approximations only holding locally). The authors should reflect on how these assumptions might be violated in practice and what the implications would be.
        \item The authors should reflect on the scope of the claims made, e.g., if the approach was only tested on a few datasets or with a few runs. In general, empirical results often depend on implicit assumptions, which should be articulated.
        \item The authors should reflect on the factors that influence the performance of the approach. For example, a facial recognition algorithm may perform poorly when image resolution is low or images are taken in low lighting. Or a speech-to-text system might not be used reliably to provide closed captions for online lectures because it fails to handle technical jargon.
        \item The authors should discuss the computational efficiency of the proposed algorithms and how they scale with dataset size.
        \item If applicable, the authors should discuss possible limitations of their approach to address problems of privacy and fairness.
        \item While the authors might fear that complete honesty about limitations might be used by reviewers as grounds for rejection, a worse outcome might be that reviewers discover limitations that aren't acknowledged in the paper. The authors should use their best judgment and recognize that individual actions in favor of transparency play an important role in developing norms that preserve the integrity of the community. Reviewers will be specifically instructed to not penalize honesty concerning limitations.
    \end{itemize}

\item {\bf Theory Assumptions and Proofs}
    \item[] Question: For each theoretical result, does the paper provide the full set of assumptions and a complete (and correct) proof?
    \item[] Answer: \answerNA{}
    \item[] Justification: The paper does not include theoretical results.
    \item[] Guidelines:
    \begin{itemize}
        \item The answer NA means that the paper does not include theoretical results. 
        \item All the theorems, formulas, and proofs in the paper should be numbered and cross-referenced.
        \item All assumptions should be clearly stated or referenced in the statement of any theorems.
        \item The proofs can either appear in the main paper or the supplemental material, but if they appear in the supplemental material, the authors are encouraged to provide a short proof sketch to provide intuition. 
        \item Inversely, any informal proof provided in the core of the paper should be complemented by formal proofs provided in appendix or supplemental material.
        \item Theorems and Lemmas that the proof relies upon should be properly referenced. 
    \end{itemize}

    \item {\bf Experimental Result Reproducibility}
    \item[] Question: Does the paper fully disclose all the information needed to reproduce the main experimental results of the paper to the extent that it affects the main claims and/or conclusions of the paper (regardless of whether the code and data are provided or not)?
    \item[] Answer: \answerYes{}
    \item[] Justification: We have the code and model checkpoints ready for release. Besides, in \Cref{append:implementation}, we provide sufficient implementation details for researchers to reproduce the results.
    \item[] Guidelines:
    \begin{itemize}
        \item The answer NA means that the paper does not include experiments.
        \item If the paper includes experiments, a No answer to this question will not be perceived well by the reviewers: Making the paper reproducible is important, regardless of whether the code and data are provided or not.
        \item If the contribution is a dataset and/or model, the authors should describe the steps taken to make their results reproducible or verifiable. 
        \item Depending on the contribution, reproducibility can be accomplished in various ways. For example, if the contribution is a novel architecture, describing the architecture fully might suffice, or if the contribution is a specific model and empirical evaluation, it may be necessary to either make it possible for others to replicate the model with the same dataset, or provide access to the model. In general. releasing code and data is often one good way to accomplish this, but reproducibility can also be provided via detailed instructions for how to replicate the results, access to a hosted model (e.g., in the case of a large language model), releasing of a model checkpoint, or other means that are appropriate to the research performed.
        \item While NeurIPS does not require releasing code, the conference does require all submissions to provide some reasonable avenue for reproducibility, which may depend on the nature of the contribution. For example
        \begin{enumerate}
            \item If the contribution is primarily a new algorithm, the paper should make it clear how to reproduce that algorithm.
            \item If the contribution is primarily a new model architecture, the paper should describe the architecture clearly and fully.
            \item If the contribution is a new model (e.g., a large language model), then there should either be a way to access this model for reproducing the results or a way to reproduce the model (e.g., with an open-source dataset or instructions for how to construct the dataset).
            \item We recognize that reproducibility may be tricky in some cases, in which case authors are welcome to describe the particular way they provide for reproducibility. In the case of closed-source models, it may be that access to the model is limited in some way (e.g., to registered users), but it should be possible for other researchers to have some path to reproducing or verifying the results.
        \end{enumerate}
    \end{itemize}

\item {\bf Open access to data and code}
    \item[] Question: Does the paper provide open access to the data and code, with sufficient instructions to faithfully reproduce the main experimental results, as described in supplemental material?
    \item[] Answer: \answerYes{}
    \item[] Justification: The code and model checkpoints have been released.
    \item[] Guidelines:
    \begin{itemize}
        \item The answer NA means that paper does not include experiments requiring code.
        \item Please see the NeurIPS code and data submission guidelines (\url{https://nips.cc/public/guides/CodeSubmissionPolicy}) for more details.
        \item While we encourage the release of code and data, we understand that this might not be possible, so “No” is an acceptable answer. Papers cannot be rejected simply for not including code, unless this is central to the contribution (e.g., for a new open-source benchmark).
        \item The instructions should contain the exact command and environment needed to run to reproduce the results. See the NeurIPS code and data submission guidelines (\url{https://nips.cc/public/guides/CodeSubmissionPolicy}) for more details.
        \item The authors should provide instructions on data access and preparation, including how to access the raw data, preprocessed data, intermediate data, and generated data, etc.
        \item The authors should provide scripts to reproduce all experimental results for the new proposed method and baselines. If only a subset of experiments are reproducible, they should state which ones are omitted from the script and why.
        \item At submission time, to preserve anonymity, the authors should release anonymized versions (if applicable).
        \item Providing as much information as possible in supplemental material (appended to the paper) is recommended, but including URLs to data and code is permitted.
    \end{itemize}

\item {\bf Experimental Setting/Details}
    \item[] Question: Does the paper specify all the training and test details (e.g., data splits, hyperparameters, how they were chosen, type of optimizer, etc.) necessary to understand the results?
    \item[] Answer: \answerYes{}
    \item[] Justification: We provided full details in \Cref{sec:exp_setup}, \Cref{append:domain} and \Cref{append:baselines}.
    \item[] Guidelines:
    \begin{itemize}
        \item The answer NA means that the paper does not include experiments.
        \item The experimental setting should be presented in the core of the paper to a level of detail that is necessary to appreciate the results and make sense of them.
        \item The full details can be provided either with the code, in appendix, or as supplemental material.
    \end{itemize}

\item {\bf Experiment Statistical Significance}
    \item[] Question: Does the paper report error bars suitably and correctly defined or other appropriate information about the statistical significance of the experiments?
    \item[] Answer: \answerYes{}
    \item[] Justification: We provided the mean and standard error over several random seeds in the experimental results to demonstrate statistical significance.
    \item[] Guidelines:
    \begin{itemize}
        \item The answer NA means that the paper does not include experiments.
        \item The authors should answer "Yes" if the results are accompanied by error bars, confidence intervals, or statistical significance tests, at least for the experiments that support the main claims of the paper.
        \item The factors of variability that the error bars are capturing should be clearly stated (for example, train/test split, initialization, random drawing of some parameter, or overall run with given experimental conditions).
        \item The method for calculating the error bars should be explained (closed form formula, call to a library function, bootstrap, etc.)
        \item The assumptions made should be given (e.g., Normally distributed errors).
        \item It should be clear whether the error bar is the standard deviation or the standard error of the mean.
        \item It is OK to report 1-sigma error bars, but one should state it. The authors should preferably report a 2-sigma error bar than state that they have a 96\% CI, if the hypothesis of Normality of errors is not verified.
        \item For asymmetric distributions, the authors should be careful not to show in tables or figures symmetric error bars that would yield results that are out of range (e.g. negative error rates).
        \item If error bars are reported in tables or plots, The authors should explain in the text how they were calculated and reference the corresponding figures or tables in the text.
    \end{itemize}

\item {\bf Experiments Compute Resources}
    \item[] Question: For each experiment, does the paper provide sufficient information on the computer resources (type of compute workers, memory, time of execution) needed to reproduce the experiments?
    \item[] Answer: \answerYes{}
    \item[] Justification: We detailed the compute resources used for the experiments in \Cref{sec:compute_resource}.
    \item[] Guidelines:
    \begin{itemize}
        \item The answer NA means that the paper does not include experiments.
        \item The paper should indicate the type of compute workers CPU or GPU, internal cluster, or cloud provider, including relevant memory and storage.
        \item The paper should provide the amount of compute required for each of the individual experimental runs as well as estimate the total compute. 
        \item The paper should disclose whether the full research project required more compute than the experiments reported in the paper (e.g., preliminary or failed experiments that didn't make it into the paper). 
    \end{itemize}
    
\item {\bf Code Of Ethics}
    \item[] Question: Does the research conducted in the paper conform, in every respect, with the NeurIPS Code of Ethics \url{https://neurips.cc/public/EthicsGuidelines}?
    \item[] Answer: \answerYes{}
    \item[] Justification: The research conducted in the paper conforms, in every respect, with the NeurIPS Code of Ethics \url{https://neurips.cc/public/EthicsGuidelines}.
    \item[] Guidelines:
    \begin{itemize}
        \item The answer NA means that the authors have not reviewed the NeurIPS Code of Ethics.
        \item If the authors answer No, they should explain the special circumstances that require a deviation from the Code of Ethics.
        \item The authors should make sure to preserve anonymity (e.g., if there is a special consideration due to laws or regulations in their jurisdiction).
    \end{itemize}

\item {\bf Broader Impacts}
    \item[] Question: Does the paper discuss both potential positive societal impacts and negative societal impacts of the work performed?
    \item[] Answer: \answerYes{}
    \item[] Justification: See the conclusion in \Cref{conclusion:limit}.
    \item[] Guidelines:
    \begin{itemize}
        \item The answer NA means that there is no societal impact of the work performed.
        \item If the authors answer NA or No, they should explain why their work has no societal impact or why the paper does not address societal impact.
        \item Examples of negative societal impacts include potential malicious or unintended uses (e.g., disinformation, generating fake profiles, surveillance), fairness considerations (e.g., deployment of technologies that could make decisions that unfairly impact specific groups), privacy considerations, and security considerations.
        \item The conference expects that many papers will be foundational research and not tied to particular applications, let alone deployments. However, if there is a direct path to any negative applications, the authors should point it out. For example, it is legitimate to point out that an improvement in the quality of generative models could be used to generate deepfakes for disinformation. On the other hand, it is not needed to point out that a generic algorithm for optimizing neural networks could enable people to train models that generate Deepfakes faster.
        \item The authors should consider possible harms that could arise when the technology is being used as intended and functioning correctly, harms that could arise when the technology is being used as intended but gives incorrect results, and harms following from (intentional or unintentional) misuse of the technology.
        \item If there are negative societal impacts, the authors could also discuss possible mitigation strategies (e.g., gated release of models, providing defenses in addition to attacks, mechanisms for monitoring misuse, mechanisms to monitor how a system learns from feedback over time, improving the efficiency and accessibility of ML).
    \end{itemize}
    
\item {\bf Safeguards}
    \item[] Question: Does the paper describe safeguards that have been put in place for responsible release of data or models that have a high risk for misuse (e.g., pretrained language models, image generators, or scraped datasets)?
    \item[] Answer: \answerNA{}
    \item[] Justification: The paper poses no such risks.
    \item[] Guidelines: 
    \begin{itemize}
        \item The answer NA means that the paper poses no such risks.
        \item Released models that have a high risk for misuse or dual-use should be released with necessary safeguards to allow for controlled use of the model, for example by requiring that users adhere to usage guidelines or restrictions to access the model or implementing safety filters. 
        \item Datasets that have been scraped from the Internet could pose safety risks. The authors should describe how they avoided releasing unsafe images.
        \item We recognize that providing effective safeguards is challenging, and many papers do not require this, but we encourage authors to take this into account and make a best faith effort.
    \end{itemize}

\item {\bf Licenses for existing assets}
    \item[] Question: Are the creators or original owners of assets (e.g., code, data, models), used in the paper, properly credited and are the license and terms of use explicitly mentioned and properly respected?
    \item[] Answer: \answerYes{}
    \item[] Justification: For all the datasets and algorithm baselines used in the paper, we have cited the original papers and provided the license, copyright information, and terms of use in the package in our code repository.
    \item[] Guidelines:
    \begin{itemize}
        \item The answer NA means that the paper does not use existing assets.
        \item The authors should cite the original paper that produced the code package or dataset.
        \item The authors should state which version of the asset is used and, if possible, include a URL.
        \item The name of the license (e.g., CC-BY 4.0) should be included for each asset.
        \item For scraped data from a particular source (e.g., website), the copyright and terms of service of that source should be provided.
        \item If assets are released, the license, copyright information, and terms of use in the package should be provided. For popular datasets, \url{paperswithcode.com/datasets} has curated licenses for some datasets. Their licensing guide can help determine the license of a dataset.
        \item For existing datasets that are re-packaged, both the original license and the license of the derived asset (if it has changed) should be provided.
        \item If this information is not available online, the authors are encouraged to reach out to the asset's creators.
    \end{itemize}

\item {\bf New Assets}
    \item[] Question: Are new assets introduced in the paper well documented and is the documentation provided alongside the assets?
    \item[] Answer: \answerYes{}
    \item[] Justification: New assets introduced in the paper are well documented and the documentation is provided alongside the assets.
    \item[] Guidelines:
    \begin{itemize}
        \item The answer NA means that the paper does not release new assets.
        \item Researchers should communicate the details of the dataset/code/model as part of their submissions via structured templates. This includes details about training, license, limitations, etc. 
        \item The paper should discuss whether and how consent was obtained from people whose asset is used.
        \item At submission time, remember to anonymize your assets (if applicable). You can either create an anonymized URL or include an anonymized zip file.
    \end{itemize}

\item {\bf Crowdsourcing and Research with Human Subjects}
    \item[] Question: For crowdsourcing experiments and research with human subjects, does the paper include the full text of instructions given to participants and screenshots, if applicable, as well as details about compensation (if any)? 
    \item[] Answer: \answerNA{}
    \item[] Justification: The paper does not involve crowdsourcing or research with human subjects.
    \item[] Guidelines:
    \begin{itemize}
        \item The answer NA means that the paper does not involve crowdsourcing nor research with human subjects.
        \item Including this information in the supplemental material is fine, but if the main contribution of the paper involves human subjects, then as much detail as possible should be included in the main paper. 
        \item According to the NeurIPS Code of Ethics, workers involved in data collection, curation, or other labor should be paid at least the minimum wage in the country of the data collector. 
    \end{itemize}

\item {\bf Institutional Review Board (IRB) Approvals or Equivalent for Research with Human Subjects}
    \item[] Question: Does the paper describe potential risks incurred by study participants, whether such risks were disclosed to the subjects, and whether Institutional Review Board (IRB) approvals (or an equivalent approval/review based on the requirements of your country or institution) were obtained?
    \item[] Answer: \answerNA{}
    \item[] Justification: The paper does not involve crowdsourcing or research with human subjects.
    \item[] Guidelines:
    \begin{itemize}
        \item The answer NA means that the paper does not involve crowdsourcing nor research with human subjects.
        \item Depending on the country in which research is conducted, IRB approval (or equivalent) may be required for any human subjects research. If you obtained IRB approval, you should clearly state this in the paper. 
        \item We recognize that the procedures for this may vary significantly between institutions and locations, and we expect authors to adhere to the NeurIPS Code of Ethics and the guidelines for their institution. 
        \item For initial submissions, do not include any information that would break anonymity (if applicable), such as the institution conducting the review.
    \end{itemize}

\end{enumerate}

\end{document}